\documentclass[10pt,twocolumn,letterpaper]{article}

\usepackage[pagenumbers]{cvpr} % To force page numbers, e.g. for an arXiv version

\definecolor{cvprblue}{rgb}{0.21,0.49,0.74}
\usepackage[pagebackref,breaklinks,colorlinks,allcolors=cvprblue]{hyperref}

\usepackage{multicol}
\usepackage{graphicx}
\usepackage{subcaption}
\usepackage{float}
\usepackage{multirow}
\usepackage{array}
\newcommand{\PreserveBackslash}[1]{\let\temp=\\#1\let\\=\temp}
\newcolumntype{C}[1]{>{\PreserveBackslash\centering}p{#1}}
\newcolumntype{R}[1]{>{\PreserveBackslash\raggedleft}p{#1}}
\newcolumntype{L}[1]{>{\PreserveBackslash\raggedright}p{#1}}
\usepackage{diagbox}
\usepackage{booktabs}
\usepackage{graphicx}
\usepackage{wrapfig}
\usepackage{caption}
\usepackage[table]{xcolor} 
\usepackage{multirow}
\usepackage{bbm}
\usepackage{enumitem}
\usepackage{amsmath,amsfonts,bm}
\usepackage{makecell}
\usepackage{amssymb}
\usepackage{tabularx}
\usepackage[capitalize]{cleveref}
\crefname{figure}{Figure}{Figure}
\crefname{table}{Table}{Table}
\crefname{section}{Section}{Section}
\crefname{appendix}{Appendix}{Appendix}

\definecolor{darkgreen}{rgb}{0,0.7,0}

\definecolor{mygraytext}{gray}{.75}

\definecolor{mygray}{gray}{.9}
\definecolor{goldenrod}{RGB}{245,245,220}
\newlength\savewidth\newcommand\shline{\noalign{\global\savewidth\arrayrulewidth\global\arrayrulewidth 1pt}\hline\noalign{\global\arrayrulewidth\savewidth}}
\newcolumntype{a}{>{\columncolor{mygray}}c}
\definecolor{mygray}{gray}{.92}
\definecolor{ForestGreen}{RGB}{34,139,34}
\newcommand{\fg}[1]{\mathbf{\mathcolor{ForestGreen}{#1}}}
\definecolor{Forestred}{RGB}{220,50,50}

\definecolor{Orange}{RGB}{255,127,0}

\let\cite\citep

\def\confName{CVPR}
\def\confYear{2026}

\title{When Token Pruning is Worse than Random: Understanding Visual Token Information in VLLMs
}

\author{
    {\bf Yahong Wang}$^{1,*}$ \quad
    {\bf Juncheng Wu}$^{2,3,*}$ \quad
    {\bf Zhangkai Ni}$^{1,\dagger}$ \quad
    {\bf Longzhen Yang}$^{1}$ \quad
    {\bf Yihang Liu}$^{1}$ \\
    {\bf Chengmei Yang}$^{1}$ \quad
    {\bf Ying Wen}$^{4}$ \quad
    {\bf Lianghua He}$^{1,5,\dagger}$ \quad
    {\bf Xianfeng Tang} \quad
    {\bf Hui Liu}$^{3}$ \quad
    {\bf Yuyin Zhou}$^{2}$ \\
    \textsuperscript{1}Tongji University \quad
    \textsuperscript{2}University of California, Santa Cruz  \quad 
    \textsuperscript{3}Amazon \\
    \textsuperscript{4}East China Normal University \quad 
    \textsuperscript{5}Shanghai Eye Disease Prevention and Treatment Center \\
}

\begin{document}
\maketitle
{
\let\thefootnote\relax
\footnotetext{* Equal contribution}
\footnotetext{
$\dagger$ Corresponding authors: 
% \mbox{\texttt{\{zkni,helianghua\}@tongji.edu.cn}}
\mbox{\scriptsize \texttt{\{zkni, helianghua\}@tongji.edu.cn}}
}

\begin{figure*}[ht]
\centering
    \includegraphics[width=0.95\textwidth]{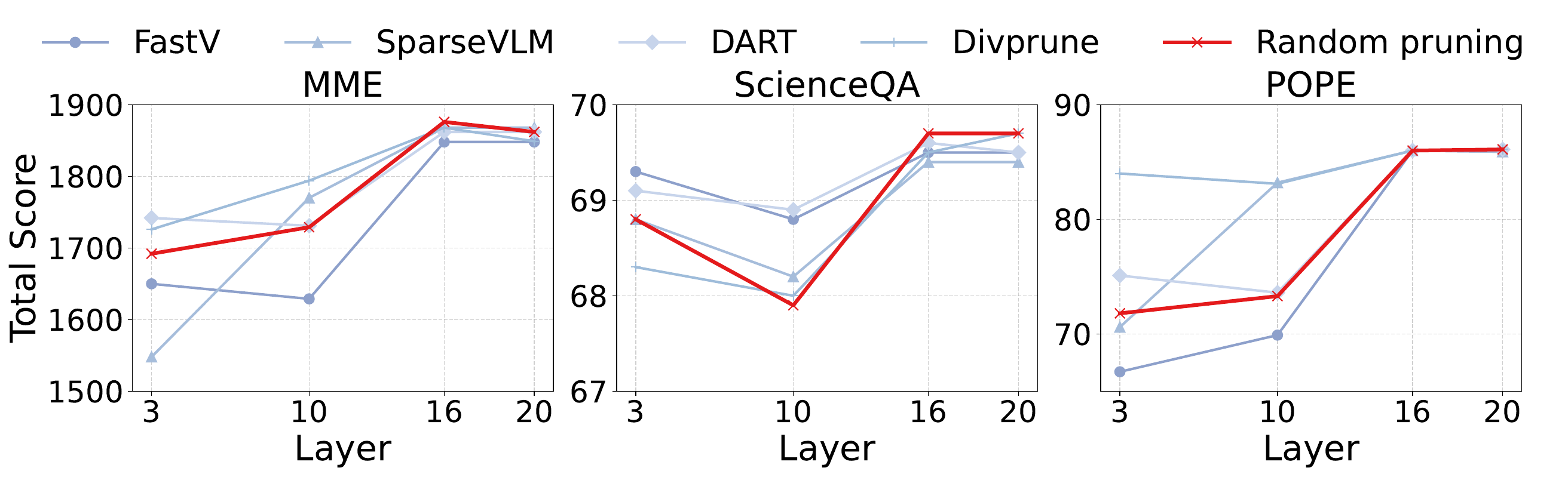}
    \caption{
    \textbf{Existing token pruning methods exhibit similar performance to random pruning at deeper layers.} We compare various pruning methods on LLaVA-1.5-7B model and three benchmarks, with 90\% of visual tokens are pruned within a given language decoder layer.
    }
    \vspace{-2em}
    \label{fig:random_vs_baselines}
\end{figure*}

\begin{abstract}

Vision Large Language Models (VLLMs) incur high computational costs due to their reliance on hundreds of visual tokens to represent images. While token pruning offers a promising solution for accelerating inference, this paper, however, identifies a key observation: in deeper layers (e.g., beyond the 20th), existing training-free pruning methods perform no better than random pruning. We hypothesize that this degradation is caused by \textbf{``vanishing token information''}, where visual tokens progressively lose their salience with increasing network depth. 
To validate this hypothesis, we quantify a token's information content by measuring the change in the model output probabilities upon its removal. 
Using this proposed metric, our analysis of the information of visual tokens across layers reveals three key findings: (1) As layers deepen, the information of visual tokens gradually becomes uniform and eventually vanishes at an intermediate layer, which we term as ``information horizon", beyond which the visual tokens become redundant;
(2) The position of this horizon is not static; it extends deeper for visually intensive tasks, such as Optical Character Recognition (OCR), compared to more general tasks like Visual Question Answering (VQA);
(3) This horizon is also strongly correlated with model capacity, as stronger VLLMs (e.g., Qwen2.5-VL) employ deeper visual tokens than weaker models (e.g., LLaVA-1.5).
Based on our findings, we show that simple random pruning in deep layers efficiently balances performance and efficiency. Moreover, integrating random pruning consistently enhances existing methods. 
Using DivPrune with random pruning achieves state-of-the-art results, maintaining 96.9\% of Qwen-2.5-VL-7B performance while pruning 50\% of visual tokens.
% Using DART with random pruning achieves state-of-the-art results, maintaining 93.9\% of Qwen-2.5-VL-7B performance while pruning 50\% of visual tokens. 
The code is available at \url{https://github.com/YahongWang1/Information-Horizon}.

\end{abstract}

\begin{figure}[ht]
\centering
    \includegraphics[width=\linewidth]{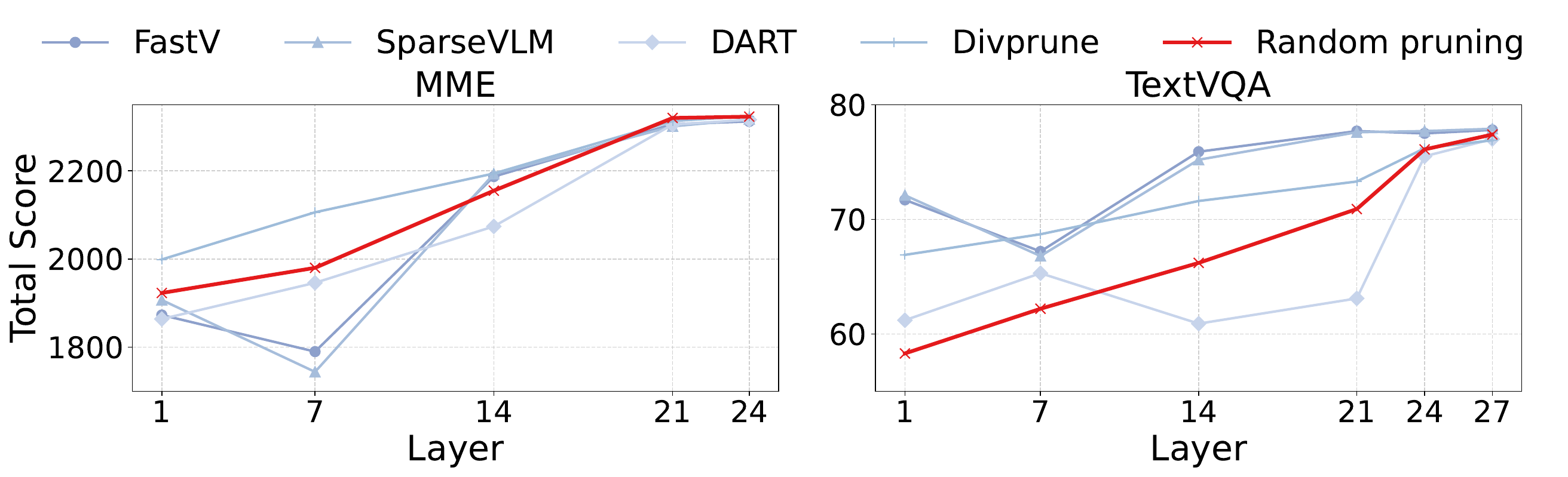}
    \caption{
    We compare various pruning methods on Qwen-2.5-VL-7B model using the MME and TextVQA benchmarks. At each decoder layer, 87.5\% of the visual tokens are removed.
    }
    \vspace{-1em}
    \label{fig:random_vs_baselines_qwen}
\end{figure}

\section{Introduction}

Vision Large Language models (VLLMs)~\citep{bai2025qwen2,li2024llava,chen2025sft,liu2023visual,zhu2023minigpt,chen2024internvl} have achieved remarkable success on a wide range of multi-modal tasks including knowledge QA~\citep{lu2022learn, fu2024mmecomprehensiveevaluationbenchmark,hudson2018gqa}, spatial reasoning~\citep{rajabi2024gsr,zhang2024mme,cheng2024spatialrgpt} and OCR~\citep{singh2019towards,mishra2019ocr,liu2024ocrbench}, by integrating a visual encoder~\cite{radford2021learning,zhai2023sigmoid, chen2024internvl} with the large language model~\cite{vicuna2023,touvron2023llamaopenefficientfoundation,2023internlm,bai2023qwentechnicalreport}. However, VLLMs usually convert the image into extensive visual tokens, which dominate the input sequence length and significantly slow down the inference process.

Reducing the number of visual tokens is therefore critical for efficient VLLM deployment, and various training-free token pruning strategies have been explored, which can be classified into two main categories according to their criteria for selecting visual tokens for pruning: (1) importance-based methods, which determine the token importance via attention weights and discard tokens deemed less important~\citep{chen2024image,zhang2024sparsevlm,ye2025fit,liu2024multi,zhao2025stitch}; (2) diversity-based methods, which assess the similarity between tokens and remove the redundant ones~\citep{bolya2022tome,alvar2025divprune,wen2025stop,wang2025folder}.
These techniques exhibit promising performance in significantly reducing inference cost while maintaining the majority of VLLMs' performance.
However, we observe a notable limitation: in deep layers of VLLMs' language decoder, existing token pruning methods perform \textbf{similarly or even worse than random pruning} (\textit{i.e.}, randomly selecting tokens to remove). 
% As shown in \cref{fig:random_vs_baselines}, we employ various token pruning methods at different layers of the LLaVA-1.5-7B model~\citep{liu2024improved}. In the deeper layers (\textit{e.g.}, 16th to 20th), none of the evaluated pruning methods show better performance than random pruning across three benchmarks~\citep{fu2024mmecomprehensiveevaluationbenchmark,lu2022learn,li2023evaluating}. 
% The corresponding results for Qwen-2.5-VL-7B~\citep{bai2025qwen2} are provided in the \cref{app:token_randomvssota}.
As shown in \cref{fig:random_vs_baselines} and \cref{fig:random_vs_baselines_qwen}, we employ various token pruning methods at different layers of the LLaVA-1.5-7B model~\citep{liu2024improved} and Qwen-2.5-VL-7B~\citep{bai2025qwen2}. 
On LLaVA-1.5-7B, none of the evaluated pruning methods show better performance than random pruning in the deeper layers (\textit{e.g.}, 16th to 20th), across three benchmarks~\citep{fu2024mmecomprehensiveevaluationbenchmark,lu2022learn,li2023evaluating}. 
A similar trend is observed on Qwen-2.5-VL-7B, where random pruning becomes competitive after the 21st layer on MME~\citep{fu2024mmecomprehensiveevaluationbenchmark} and after the 24th layer on TextVQA~\citep{singh2019towards}.
More results are provided in the \cref{app:token_randomvssota}.
These findings lead to an intriguing question: \textit{when performing no better than random, can these pruning methods identify visual tokens containing information necessary to produce the answer?}

To answer this question, we firstly propose to estimate a visual token's information by measuring changes in model output probabilities when this token is removed. 
Specifically, as illustrated in \cref{fig:teaser}, we initially prune all visual tokens except the target one at a specific layer and calculate the model's predicted probability on the ground-truth label.
Subsequently, we further remove this visual token at the same layer, forcing the model to rely solely on text tokens. Finally, the difference in probabilities with and without the target visual token serves as the estimate of its information at the specified layer.
% Both theoretical and experimental evidences are provided to illustrate the effectiveness of the proposed information modeling: (a) our definition is aligned with the information bottleneck (IB) principle; and (b) removing low-information tokens based on our definition consistently improves the model's performance.
Experimental evidence illustrates the effectiveness of the proposed information modeling: removing low-information tokens based on our definition consistently improves the model's performance.

\begin{figure*}[t]
    \includegraphics[width=1.0\textwidth]{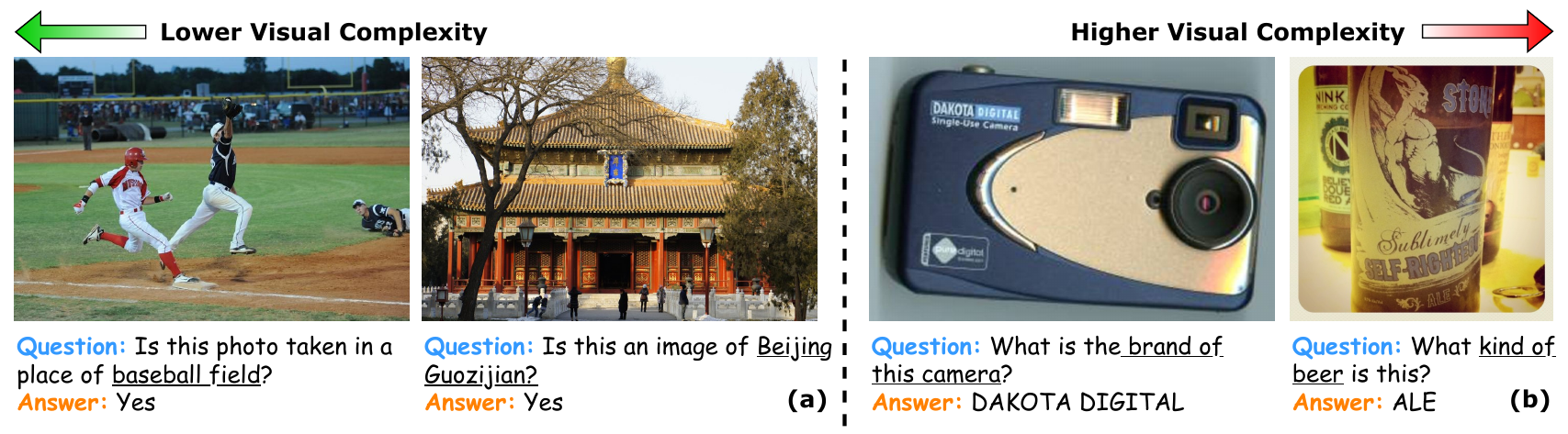}
    \caption{
    \textbf{Tasks with different visual complexity.} Low visual complexity tasks only require the VLLM to identify global information such as the main scene, while tasks with higher visual complexity require the model to focus on visual details.}
    \label{fig:visual_difficulty}
\end{figure*}

From the lens of visual token information, we rethink the failure of existing token pruning methods in deep layers.
Our experimental results demonstrate that: most pruning techniques effectively retain visual tokens with high information up to the 10th layer of the LLaVA-1.5-7B model. Beyond the 14th layer, however, random pruning similarly maintains visual token information.
The primary cause is that visual token information vanishes uniformly as layers deepen, ultimately reaching zero beyond a specific layer (see \cref{fig:imp_varwithheatmap}), which we term as ``\textit{information horizon}".
In fact, the visual tokens after the information horizon could be entirely removed without compromising the model's performance.
We further illustrate that the position of the information horizon is dynamic, influenced by two primary factors.
(1) Task visual complexity: comparing to tasks where the model simply answers knowledge questions based on objects~\citep{fu2024mmecomprehensiveevaluationbenchmark,lu2022learn} (see \cref{fig:visual_difficulty}(a)), the tasks demanding more visual cues~\citep{singh2019towards,mishra2019ocr,liu2024ocrbench}, as shown \cref{fig:visual_difficulty}(b), depend on deeper layer visual tokens; (2) Model visual capability: for the same task, the models with superior visual abilities (\textit{e.g.}, Qwen-2.5-VL-7B~\citep{bai2025qwen2}) can exploit deeper layers visual tokens than less advanced models like LLaVA-1.5-7B~\citep{liu2024improved}, extending its information horizon to deeper layers.

Due to the dynamic information horizon, removing all visual tokens at a fixed layer could hinder performance in tasks with higher visual complexity or for models with enhanced visual abilities.
In this paper, we demonstrate that simply integrating random pruning with existing pruning methods can more effectively balance inference efficiency and accuracy across various datasets. 
For example, in the case of Qwen-2.5-VL-7B, combining random pruning with DART enhances performance on OCRBench from 75.5\% to 77.9\%, maintaining 93.9\% of the initial model performance while removing 50\% visual tokens. 
For LLaVA-1.5-7B, employing DivPrune + Random pruning results in a 6.7\% improvement on MMBench compared to using only DivPrune (61.3\% \textit{vs.} 54.6\%). 
Alternatively, arming LLaVA with DART + Random pruning, the CUDA latency and FLOPs can decrease by 73.0\% and 74.4\%, while maintaining 91.6\% of its original performance.
Our contributions are summarized as follows: 
\begin{itemize}
    \item We propose to quantify the visual token information in VLLMs by calculating the change in output probabilities, demonstrating that removing low-information visual tokens improves the model's performance.

    \item We observe that the information in visual tokens gradually becomes uniform and eventually disappears at an intermediate layer (information horizon). Post this layer, visual tokens can be discarded without affecting model performance.
    
    \item We demonstrate that both task visual complexity and model visual capability impact the position of information horizon, illustrating that integrating random pruning with existing pruning methods could more effectively balance the performance and efficiency.
\end{itemize}

\section{Related Work}
\paragraph{Visual Large Language Model.}
% Visual Large Language Models (VLLMs) demonstrate impressive capabilities, showing strong performance on diverse multimodal tasks.
% Visual Large Language Models (VLLMs) demonstrate impressive capabilities, showing strong performance on diverse multimodal tasks including (i) perception \& grounding (\textit{e.g.}, MME~\citep{fu2024mmecomprehensiveevaluationbenchmark}), (ii) knowledge-intensive multimodal reasoning (\textit{e.g.}, ScienceQA~\citep{lu2022learn}), and (iii) text-centric understanding that requires reading in-image text (\textit{e.g.}, TextVQA~\citep{singh2019towards} and OCRBench~\citep{liu2024ocrbench}).
Visual Large Language Models (VLLMs) demonstrate impressive capabilities, showing strong performance on diverse multimodal tasks~\cite{huang2025medvlthinker,li2024llava} including (i) perception \& grounding~\citep{fu2024mmecomprehensiveevaluationbenchmark,ma2024groma,li2023evaluating}, (ii) knowledge-intensive multimodal reasoning~\citep{lu2022learn,hudson2018gqa,fu2025livevqa}, and (iii) text-centric understanding that requires reading in-image text~\citep{singh2019towards,liu2024ocrbench,mathew2021docvqa}.
% A typical VLLM consists of three main components: (i) a vision encoder that extracts visual features, (ii) a vision--language connector that maps visual embeddings into language feature space, and (iii) a pretrained large language model that performs reasoning and generation.
A typical VLLM comprises three components: (i) a vision encoder, (ii) a vision–language connector, and (iii) a pretrained large language model for reasoning and generation.
While this architecture has become the dominant paradigm, it incurs high computational costs as VLLMs often encode images into numerous tokens. For example, LLaVA-1.5 encodes a $336 \times 336$ image into 576 tokens~\cite{liu2024improved}, while Qwen2.5-VL can produce thousands of tokens for high-resolution images~\cite{bai2025qwen2}. Long visual sequences introduce substantial computational overhead, urging further research on sparsification to reduce redundancy while preserving essential information for downstream tasks.

\paragraph{Visual Token Pruning.}
To mitigate the computational overhead introduced by long visual sequences, recent works explore \emph{visual token pruning} as an efficient acceleration strategy. Existing training-free visual token pruning methods can be broadly divided into two categories: \emph{importance-based} and \emph{diversity-based}. \textbf{Importance-based methods} rely on task-driven importance signals, typically guided by attention~\citep{chen2024image,zhang2024sparsevlm,zhao2025stitch,ju2024turbo,xing2024pyramiddrop}. FastV~\cite{chen2024image} exploits the attention of the final text token, which directly determines the next output token, as a principled signal to eliminate redundant visual tokens, while SparseVLM~\cite{zhang2024sparsevlm} adopts a two-stage strategy by first selecting salient text tokens and subsequently leveraging their attention to more precisely guide visual token pruning. SGL~\cite{zhao2025stitch} introduces a small-to-large guidance paradigm, where global attention aggregated from a small VLLM is used to guide token pruning in a large VLLM. PDrop~\citep{xing2024pyramiddrop} performs pruning progressively across layers leverages the attention of the final text token. Despite their effectiveness, importance-based methods inherently rely on attention maps, making them incompatible with FlashAttention~\cite{dao2022flashattention, dao2023flashattention2}. \textbf{Diversity-based methods} typically calculates similarity between visual tokens and removes redundant tokens, offering efficient acceleration compatibility~\citep{bolya2022tome,alvar2025divprune,wen2025stop,zhang2025cdpruner,li2025todrevisualtokenpruning}. For example, DART~\cite{wen2025stop} prunes redundant tokens by explicitly measuring duplication, and DivPrune~\cite{alvar2025divprune} formulates token pruning as a Max–Min Diversity Problem to maximize the diversity of retained tokens. Although both categories strike a reasonable balance between inference speed and accuracy, we observe that, as depicted in \cref{fig:random_vs_baselines}, pruning at deeper layers can lead to performance similar to random pruning. This motivates our investigation into the role and characteristics of visual tokens at deep layers.

\section{Methodology}

\begin{figure*}[t]
    \centering
    \includegraphics[width=\linewidth]{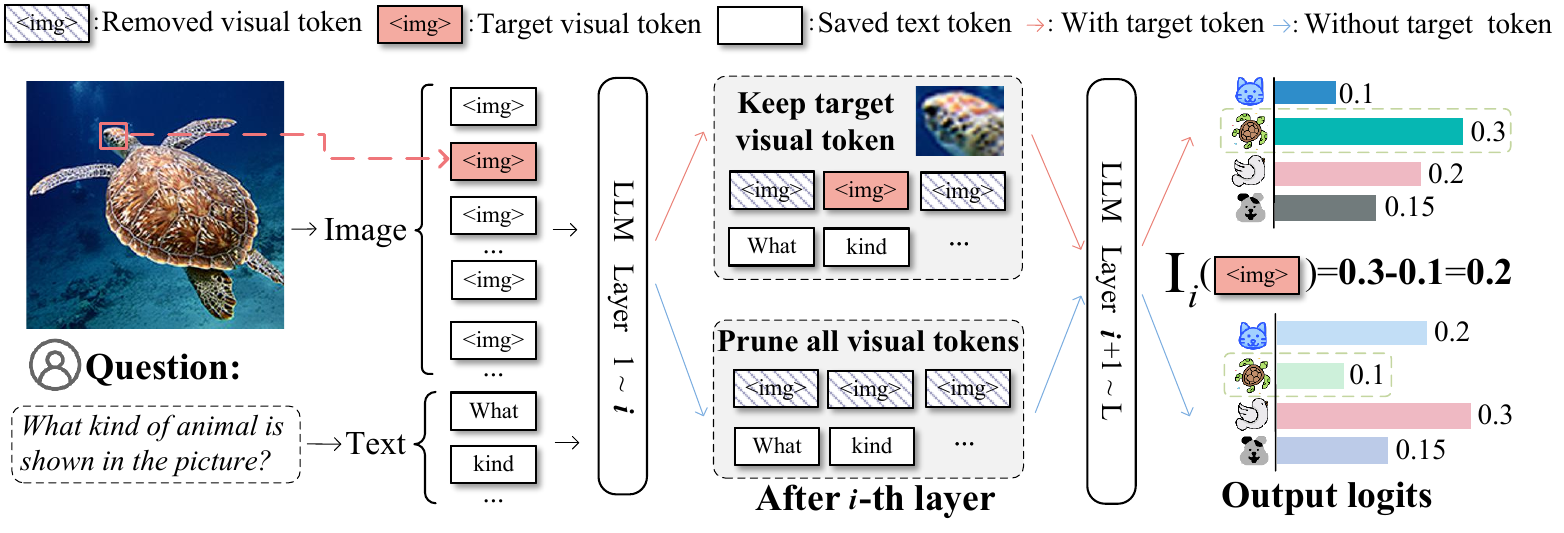}
    \caption{
    \textbf{Illustration of our framework for computing visual token information}. 
    At the $i$-th layer of VLLM's language decoder, we firstly remove all other visual tokens except the target one and run one forward pass. Next we additionally run another forward pass by further removing the only one visual token. The difference between these two output probabilities on the ground-truth label defines the information score $\text{I}_i(\mathbf{V}_k)$. 
    }
    \vspace{-1em}
    \label{fig:teaser}
\end{figure*}

In this section, we first provide an overview of VLLM in \cref{sec:preliminary}, and introduce the definition of \textbf{visual token information} in \cref{sec:token_information}. 
% Subsequently, \cref{sec:information_effectiveness} demonstrates the effectiveness of the proposed information modeling via theoretical and experimental evidence.
Subsequently, \cref{sec:information_effectiveness} demonstrates the effectiveness of the proposed information modeling via experimental evidence.

\subsection{Preliminary: Vision Large Language Model}
\label{sec:preliminary}
% \wjc{Yahong TODO: check the annotation in this section to align with definition of visual token information in \cref{sec:token_information}}

% Current mainstream architecture of LVLM follow similar design paradigms: a visual encoder, a modality projector and a language decoder. For an input image $\mathbf{X}_v$, the pretrained visual encoder like CLIP's vision model, is utilized to extract the visual feature $ \mathbf{Z}_v = g(\mathbf{X}_v) $, where $g(.)$ represents the visual encoder. To bridge the modality gap between vision and language, the projector module converts the visual features  $ \mathbf{Z}_v $ into language embedding tokens $ \mathbf{H}_v $. The language decoder $f_\theta(.)$ then integrates the visual and text tokens to generate the response in an auto-regressive manner:

% \[
% p(\mathbf{X}_G^i)=f_\theta(\mathbf{X}_v,\mathbf{X}_{instruct},\mathbf{X}_G^{1:i-1})
% \]

% where $p(\mathbf{X}_G^i)$ denotes the probability distribution of the  $i$-th generated token $\mathbf{X}_G^i$, and $\mathbf{X}_{instruct}$ is the associated text inputs.

Current mainstream architectures of Vision Large Language Models (VLLMs) follow a common design paradigm, consisting of a visual encoder, a modality projector, and a language decoder. Given an input image $\mathbf{X}_v$, a pretrained visual encoder (\textit{e.g.}, CLIP~\cite{radford2021learning} or SigLIP~\cite{zhai2023sigmoid} vision encoder) is used to extract image features $\mathbf{Z}_v = g(\mathbf{X}_v)$, where $g(\cdot)$ denotes the visual encoder. To align visual and language modalities, a projection module maps $\mathbf{Z}_v$ into visual embeddings $\mathbf{V} \in \mathbb{R}^{d\times N_v}$, where $d$ denotes the embedding dim and $N_v$ is the number of visual tokens, which is fixed as 576 in LLaVA-1.5~\cite{liu2024improved} or scales based on image resolution in Qwen2.5-VL~\cite{bai2025qwen2}.
Subsequently, $\mathbf{V}$ is fed into a pre-trained language decoder $f_\theta(\cdot)$, which integrates both visual and textual embeddings to predict the probability of the next generated token in an auto-regressive manner:
\begin{equation}  
p(g_i) = f_\theta(\mathbf{V}, \mathbf{T}, \mathbf{G}^{1:i-1}), \quad
g_i = \arg\max_{x} \, p(x)
\end{equation}  
where $\mathbf{T}\in \mathbb{R}^{d\times N_t}$ represents the embeddings of input textual tokens (\textit{e.g.}, instruction and question), $g_i$ is the $i$-th generated token, and $p(g_i)$ denotes its probability distribution over the vocabulary. $\mathbf{G}^{1:i-1} \in \mathbb{R}^{(i-1)\times d}$ represents the embeddings of previously generated tokens.
At each decoding step, the token with the highest predicted probability is appended to the generated sequence, which is then used as input for the next step. 
Throughout this section, we concentrate on the \emph{prefill} step, which predicts the \textit{first output token} probability as $p(g_1)=f_\theta(\mathbf{V}, \mathbf{T})$.

\subsection{Visual Token Information}
\label{sec:token_information}

% \begin{figure}[t]
%     \centering
%     \begin{subfigure}[b]{0.49\textwidth}
%         \centering
%         \includegraphics[width=\textwidth]{fig/information_prune/imp_prune_mme_purple.pdf}
%         \vspace{-6mm}
%         \caption{\scriptsize MME}
%         \label{fig:mme}
%     \end{subfigure}
%     \begin{subfigure}[b]{0.49\textwidth}
%         \centering
%         \includegraphics[width=\textwidth]{fig/information_prune/imp_prune_textvqa.pdf}
%         \vspace{-6mm}
%         \caption{\scriptsize TextVQA}
%         \label{fig:textvqa}
%     \end{subfigure}
%     \caption{
%     \textbf{Effectiveness of proposed visual token information modeling.}
%     Performance of LLaVA-1.5-7B on MME (a) and TextVQA (b) when pruning 88\% and 75\% visual tokens with low information at different layers.
%     Layer 0 denotes pruning applied between the visual encoder and the language decoder. 
%     Notably, removing the low-information visual tokens consistently improves model's performance across different benchmarks.
%     }
%     \label{fig:imp_prune_result}
% \end{figure}

% \wjc{Yahong TODO: define the visual token information using equation here.}

Let $\mathcal{V} = \{1,\dots, N_v\}$ denote the set of indices for the visual tokens, $y \in \{1, \dots, |V|\}$ represent the vocabulary index of the first generated token in the ground-truth answer, where $|V|$ is the length of the vocabulary. 
For a specific visual token index $k \in \mathcal{V}$, we denote the information contributed by $\mathbf{V}_k$ at the $i$-th layer of  VLLM's decoder as $\text{I}_i(\mathbf{V}_k)$.
As illustrated in Figure~\ref{fig:teaser}, to estimate $\text{I}_i(\mathbf{V}_k)$, we measure the change in the probability of the first generated token $g_1$ on $y$ when $\mathbf{V}_k$ is removed at the $i$-th layer.

Specifically, let $\mathbf{H}_v^i, \mathbf{H}_t^i = f_\theta^{1:i}(\mathbf{V}, \mathbf{T})$ denote the hidden states of the visual and text embeddings after the first $i$ layers of the decoder $f_\theta(\cdot)$, and
$\mathbf{H}_v^i = \{ \mathbf{H}_1^i, \mathbf{H}_2^i, \dots, \mathbf{H}_{N_v}^i\}$.
To isolate the interference from other visual tokens, we first  employ a binary mask $\mathbf{M} \in \{0,1\}^{1 \times N_v}$ to remove all visual tokens except $\mathbf{H}_k^i$:
\begin{equation}  
M_j = 
\begin{cases} 
1 & \text{if } j = k, \\
0 & \text{otherwise}.
\end{cases}
\end{equation}  
The masked visual representation is $\hat{\mathbf{H}}_v^i = \mathbf{H}_v^i \odot \mathbf{M}$, where $\odot$ denotes element-wise multiplication.
Next, we pass the masked hidden states through the remaining decoder layers $f_\theta^{i+1:L}$, where $L$ denotes the total number of decoder layers in the model, to yield a distribution over the vocabulary:
\begin{equation}  
p(g_1) = f_\theta^{i+1:L}(\hat{\mathbf{H}}_v^i, \mathbf{H}_t^i).
\end{equation} 
The probability assigned to the ground-truth label $y$ could be represented as:
\begin{equation}
p_k = p(g_1)[y].
\end{equation} 
Subsequently, to measure the incremental contribution of $\mathbf{V}_k$, we compare $p_k$ with the probability predicted using no visual information after the $i$-th layer. Specifically, we set $\hat{\mathbf{H}}_v^i = \mathbf{0}$ to force the model to rely solely on text hidden states:
\begin{equation}
p_{\text{text}} = f_\theta^{i+1:L}(\mathbf{0}, \mathbf{H}_t^i)[y].
\end{equation} 
Finally, the information attributed to token $\mathbf{X}_k$ at layer $i$ is defined as:
\begin{equation}
\text{I}_i(\mathbf{V}_k) = p_k - p_{\text{text}},
\label{info_def}
\end{equation} 
% as illustrated in Figure~\ref{fig:teaser}.

% We propose a method to compute the information of the visual token by measuring the individual contribution of each token to the model output. Specifically, in a chosen intermediate layer of the information (\textit{e.g.} right before the language decoder), we remove all visual tokens except the token of interest, effectively isolating the contribution of that token. We then forward-propagate only this single visual token (with the textual input intact) and record the model’s output logit corresponding to the ground-truth answer. This logit value serves as the information for the visual token, as illustrated in Figure \ref{fig:teaser}. 

% \begin{figure*}
% \centering
% \includegraphics[width=0.95\textwidth]{fig/information_prune/imp_prune_mme_textvqa_v0924.pdf}
% \vspace{-2em}
% \caption{
% \textbf{Effectiveness of proposed visual token information modeling.}
% Performance of LLaVA-1.5-7B on MME (a) and TextVQA (b) when pruning 88\% and 75\% visual tokens with low information at different layers.
% Layer 0 denotes pruning applied between the visual encoder and the language decoder. 
% Notably, removing the low-information visual tokens consistently improves the model's performance across different benchmarks.
% }
% \label{fig:imp_prune_result}
% \end{figure*}

\begin{figure*}[t]
\centering
\includegraphics[width=\textwidth]{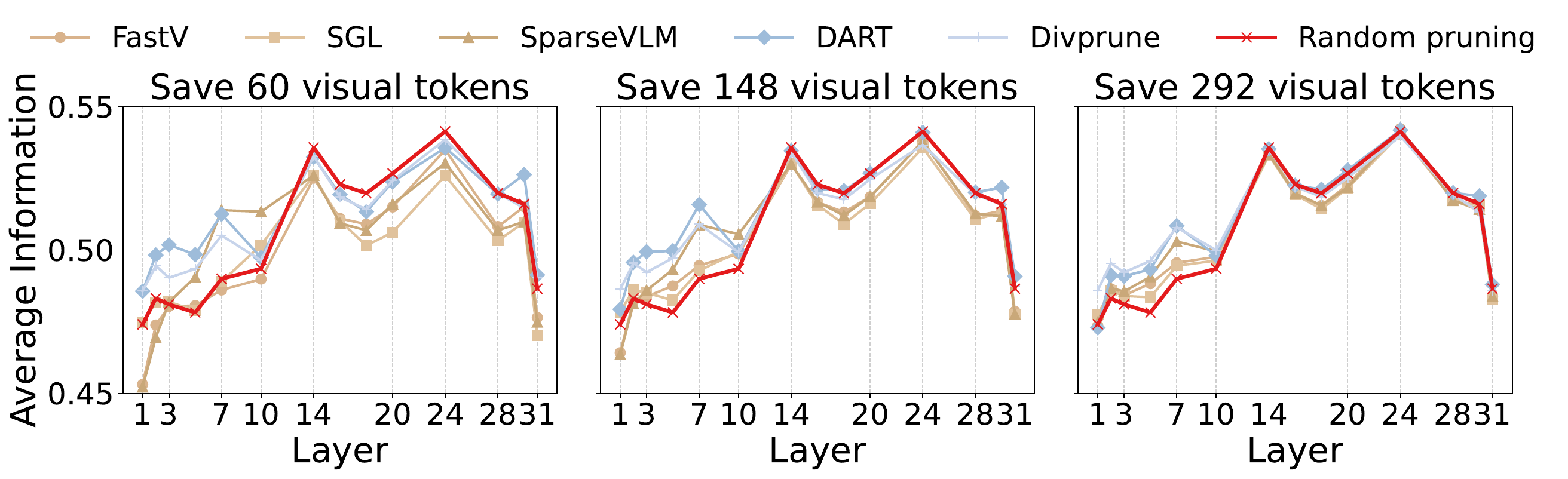}
\vspace{-3mm}
\caption{\textbf{Evaluation of various pruning methods.} We measure the sum of information in retained visual tokens when using different pruning methods.
In the deep layers, existing pruning methods fail to retain more high-information than random pruning.
}
\vspace{-3mm}
\label{fig:eval_LLM}
\end{figure*}

\begin{figure}
\centering
\includegraphics[width=0.45\textwidth]{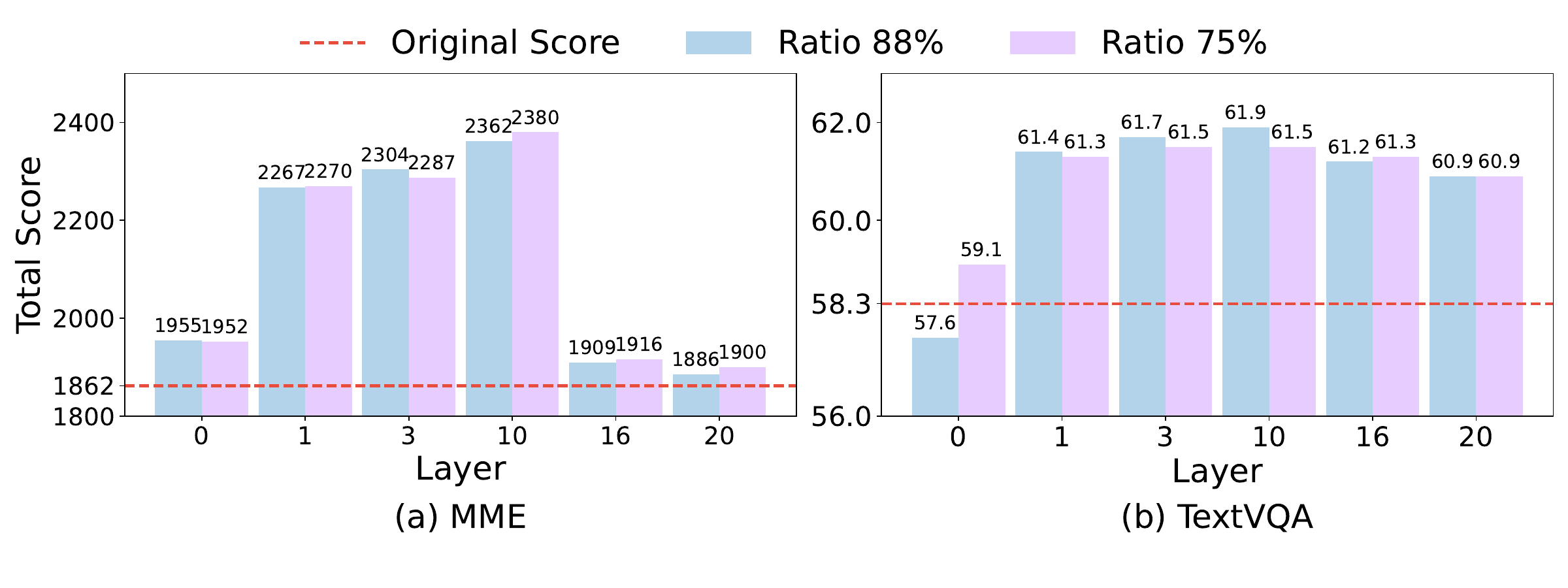}
\vspace{-1em}
\caption{
\textbf{Effectiveness of proposed visual token information modeling.}
Performance of LLaVA-1.5-7B on MME (a) and TextVQA (b) when pruning 88\% and 75\% visual tokens with low information at different layers.
Layer 0 denotes pruning applied between the visual encoder and the language decoder. 
Notably, removing the low-information visual tokens consistently improves the model's performance across different benchmarks.
}
\vspace{-2em}
\label{fig:imp_prune_result}
\end{figure}

\subsection{Effectiveness of visual token information}
\label{sec:information_effectiveness}

% \paragraph{Experimental Evidence.} 
Using the proposed visual token information, we perform extensive pruning experiments using LLaVA-1.5-7B~\citep{liu2024improved}, a widely used baseline VLLM that encodes images into a fixed sequence of 576 visual tokens.
Specifically, at different layers of VLLM's language decoder, we
rank the visual tokens based on their information defined in Equation \ref{info_def}.
Subsequently, we remove 75\% and 88\% of low-information tokens, retaining 144 and 72 visual tokens, respectively.
Our experiments are conducted on two popular VQA benchmarks: MME~\citep{fu2024mmecomprehensiveevaluationbenchmark} and TextVQA~\citep{singh2019towards}.

As illustrated in \cref{fig:imp_prune_result}, across two evaluated benchmarks, removing 75\% of low-information visual tokens consistently \textit{outperform the base model without pruning}, suggesting that our method effectively quantifies the information contributing to the ground-truth answer in the visual tokens.
Notably, our experimental results indicate that low-information visual tokens are unnecessary and may hinder inference. For example, pruning 75\% of low-information tokens at the 10th layer leads to 27.8\% and 6.1\% of performance improvement on MME and TextVQA, respectively.
We further report the corresponding results for Qwen-2.5-VL-7B in Appendix \ref{app:information_effectiveness_qwen}. 
Results on Qwen-2.5-VL-7B demonstrate the robustness of this phenomenon, where removing low-information visual tokens achieves comparable performance to the base model.
This is possibly due to low-information tokens interfering with the focus of VLLM on tokens with important information.

\noindent
% \begin{figure}
% \centering
% \includegraphics[width=\textwidth]{fig/evaluation_existing_method/tknimp_minus_systext_normal_no436_v0923.pdf}
% \caption{\textbf{Evaluation of various pruning methods.} We measure the sum of information in retained visual tokens when using different pruning methods.
% In the deep layers, existing pruning methods fail to retain more high-information than random pruning.
% }
% \label{fig:eval_LLM}
% \end{figure}

\begin{figure*}[t]
    \centering
    \includegraphics[width=\linewidth]{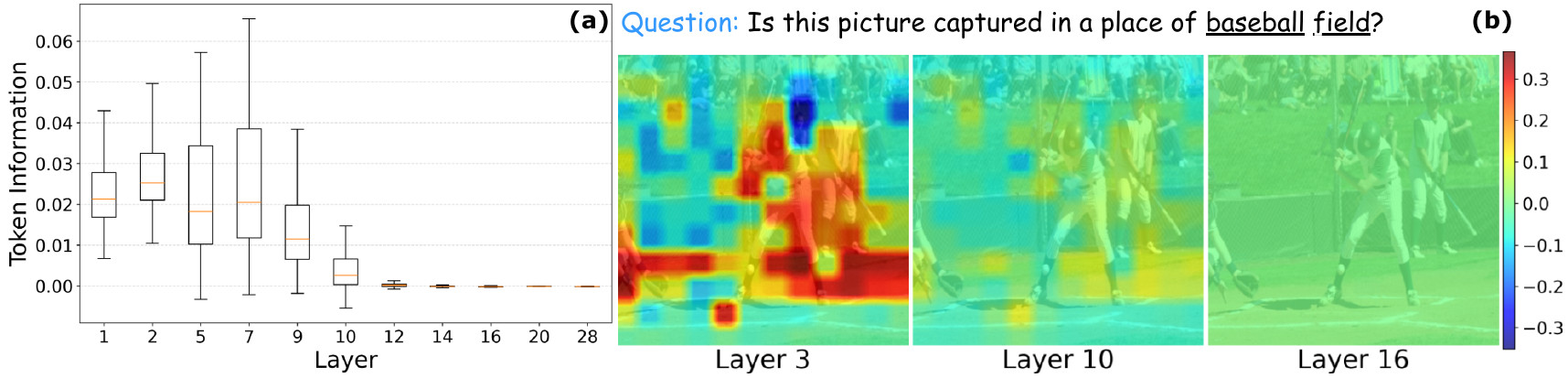}
    \vspace{-8mm}
    \caption{
        \textbf{Visual token information becomes uniform in the deep layers.}
        (a) The variance of visual token information across different layers of the language decoder.
        (b) Layer-wise visual token information maps at layers 3, 10, and 16. 
    }
    \label{fig:imp_varwithheatmap}
    \vspace{-6mm}
\end{figure*}

\begin{figure}[t]
    \centering
    \includegraphics[width=\linewidth]{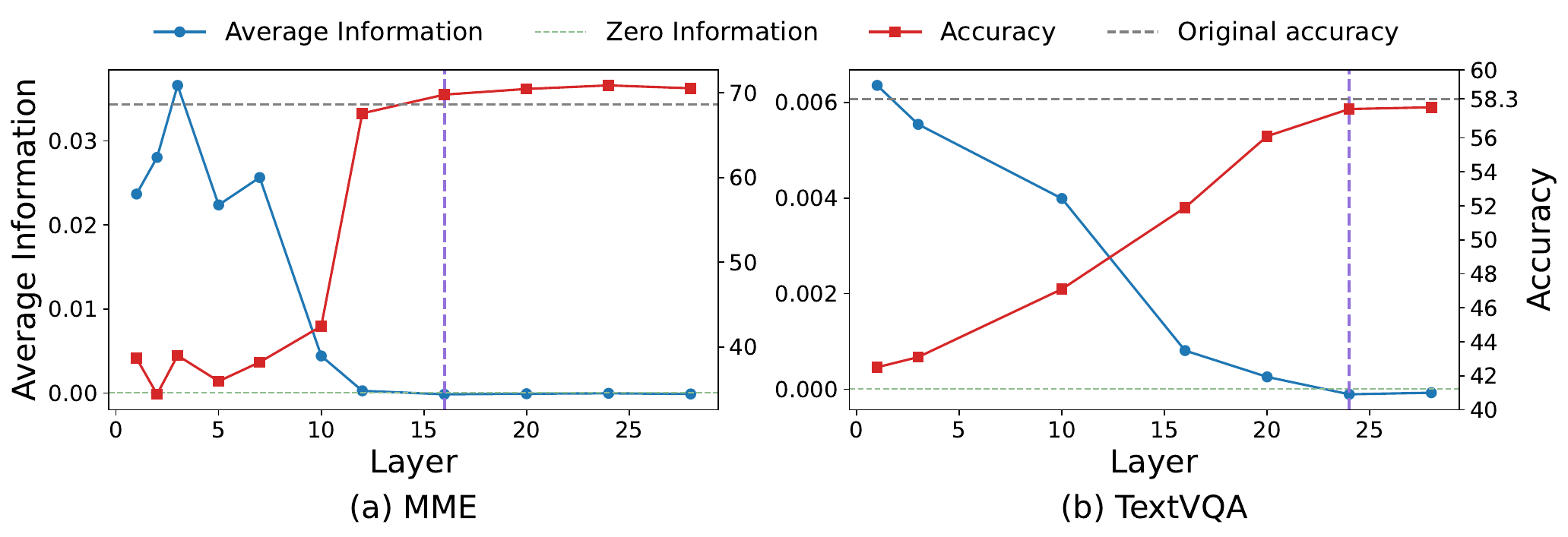}
    \vspace{-8mm}
    \caption{
    \textbf{Information horizon of visual tokens.} 
    The purple vertical dashed line marks the ``information horizon".
    When the mean information of all visual tokens becomes close to zero at a certain layer of LLaVA-1.5-7B, removing these tokens at this layer does not impair the model's performance.
    }
    \vspace{-6mm}
    \label{fig:meaninfo_layerprune}
\end{figure}

\section{Experimental Analysis}
\subsection{Understanding token pruning using information}
\label{sec:eval}

Firstly, we aim to investigate whether current pruning strategies can preserve high-information visual tokens and understand why they may perform no better than random pruning in the deep layers.
Specifically, we employ token pruning strategies at different layers of VLLM to remove some visual tokens and measure the sum of information of the remaining ones.
Our experiments are conducted on LLaVA-1.5-7B~\cite{liu2024improved} and use 200 randomly selected samples from the MME benchmark~\cite{fu2024mmecomprehensiveevaluationbenchmark}.
We evaluate a diverse set of training-free pruning methods including both importance-based and diversity-based methods, including DivPrune \citep{alvar2025divprune}, FastV \citep{chen2024image}, SparseVLMs \citep{zhang2024sparsevlm}, DART \citep{wen2025stop}, SGL \citep{zhao2025stitch}.
For each pruning method, we adopt three pruning ratios—90\%, 75\%, and 50\%—corresponding to retaining 60, 148, and 292 visual tokens, respectively.

The evaluation results are summarized in \cref{fig:eval_LLM}. At shallow layers (\textit{e.g.}, 1st to 7th), most pruning methods can more effectively retain high-information visual tokens than random pruning.
Notably, across three pruning ratios, diversity-based methods (DivPrune and DART) consistently outperform importance-based methods (SparseVLMs, SGL, FastV), suggesting that diverse visual tokens might carry more information than high-attention-scored ones.
After the 10th layer, however, the gap between baseline methods and random pruning gradually narrows. 
By the 14th layer, random pruning surpasses all baseline methods, and this continues in the subsequent layers, consistent with accuracy trends in deeper layers (as shown in \cref{fig:random_vs_baselines}).
We further repeat the experiments on Qwen2.5-VL-7B, observing highly consistent results with those of LLaVA-1.5-7B, reinforcing that this phenomenon is not model-specific. The results are provided in the \cref{app:eval_qwen}.
The underlying cause could be that \textit{the visual tokens information uniformly vanish in the deep layers.} As shown in \cref{fig:imp_varwithheatmap}, the visual tokens capture varying amounts of information across layers 1 to 7.
High-information tokens cluster around key areas, such as the baseball player, whereas low-information tokens provide limited clues about the baseball, like the grass.
However, from the 9th layer, the variability in token information starts diminishing.
By the 16th layer, the visual tokens uniformly capture negligible information. 
In this case, the selection of pruned visual tokens does not influence model performance, resulting in similar results for existing pruning methods and random pruning.

Furthermore, when the information of all visual tokens is uniformly low at a specific layer, we can remove \textbf{all visual tokens} from this layer without affecting model performance.
We refer to this layer as the ``information horizon".
As demonstrated in \cref{fig:meaninfo_layerprune},
we measure the mean information of all visual tokens across different layers and evaluate the LLaVA-1.5-7B performance when pruning all visual tokens at each layer.
For the MME benchmark (\cref{fig:meaninfo_layerprune}(a)), at layer 16, the mean information of all visual tokens nearly becomes zero, and removing these tokens at this layer results in performance close to the original model.
This phenomenon can also be observed at the 24th layer and TextVQA benchmark (see \cref{fig:meaninfo_layerprune}(b)).

% \begin{figure}[t]
%     \centering
%     \includegraphics[width=\linewidth]{fig/token_information_trend/tknvis_v0925.pdf}
%     \caption{
%         \textbf{Visual token information becomes uniform in the deep layers.}
%         (a) The variance of visual token information across different layers of the language decoder.
%         (b) Layer-wise visual token information maps at layers 3, 10, and 16. 
%     }
%     \label{fig:imp_withheatmap}
% \end{figure}

% \begin{figure*}[t]
%     \centering
%     \includegraphics[width=\linewidth]{fig/mean_information&layer_prune/textvqamme_meaninfo_layerprune_v0925.pdf}
%     \caption{
%     \textbf{Information horizon of visual tokens.} 
%     The purple vertical dashed line marks the ``information horizon".
%     When the mean information of all visual tokens becomes close to zero at a certain layer of LLaVA-1.5-7B, removing these tokens at this layer does not impair the model's performance.
%     }
%     \vspace{-1em}
%     \label{fig:meaninfo_layerprune}
% \end{figure*}

\subsection{Position of the information horizon}
\label{sec:When_deep}
To further investigate at which layer all visual tokens can be removed without loss of model performance, we conduct comprehensive experiments on Qwen-2.5-VL-7B and LLaVA-1.5-7B across 6 benchmarks, including MME~\cite{fu2024mmecomprehensiveevaluationbenchmark}, ScienceQA~\cite{lu2022learn}, POPE~\cite{li2023evaluating}, TextVQA~\cite{singh2019towards}, OCRBench~\cite{liu2024ocrbench}, and OCRVQA~\cite{mishra2019ocr}. 
As illustrated in \cref{fig:prune_all}, we find there are two primary factors:

\vspace{-1em}

\paragraph{Task visual complexity.} Comparing to knowledge question-answering~\cite{lu2022learn} or hallucination detection~\cite{li2023evaluating}, more visually complex tasks such as OCR~\cite{liu2024ocrbench} (which depend on precise visual details) rely on deeper visual tokens.
For example, with Qwen-2.5-VL-7B (\cref{fig:prune_all}(a)), pruning all visual tokens at the 20th layer leads to the highest accuracy for ScienceQA, MME, and POPE, whereas for OCRBench and TextVQA, this occurs around the 27th layer.
A similar trend can also be observed on LLaVA-1.5-7B (15th \textit{vs.} 24th layer in \cref{fig:prune_all}(b)).

\vspace{-1em}

\paragraph{Model visual capability.}
We also compare different VLLMs under the same task. 
Models with stronger visual understanding, such as Qwen-2.5-VL-7B, are able to exploit informative visual tokens from deeper layers than weaker models like LLaVA-1.5-7B. 
For example, Qwen-2.5-VL-7B leverages the visual token at the 20th layer, enhancing question-answering precision compared to LLaVA-1.5-7B (96.4\% \textit{vs.} 68.6\% on MME), with all visual tokens removable without loss from the 16th layer.
Moreover, on TextVQA and OCRBench, Qwen-2.5-VL-7B attains approximately 80\% accuracy using visual tokens by the 27th layer, whereas LLaVA-1.5-7B remains below 40\% beyond the 24th layer. 
% \wjc{Yahong please check the numbers here}

% \begin{figure*}[t]
%     \centering
%     \includegraphics[width=\linewidth]{fig/prune_all_token/qwen_vs_llava_shared_legend_v0922.pdf}
%     \caption{\textbf{Performance of models when pruning all visual tokens at different decoder layers}. Colored vertical dashed lines mark the convergence layers for datasets of different visual complexity, beyond which pruning all visual tokens has little impact on performance. This demonstrates that the layer at which visual tokens could be entirely removed depends on task visual complexity and model visual capability.
%     }

%     \label{fig:prune_all}
% \end{figure*}

\begin{figure}[t]
    \centering
    \includegraphics[width=\linewidth]{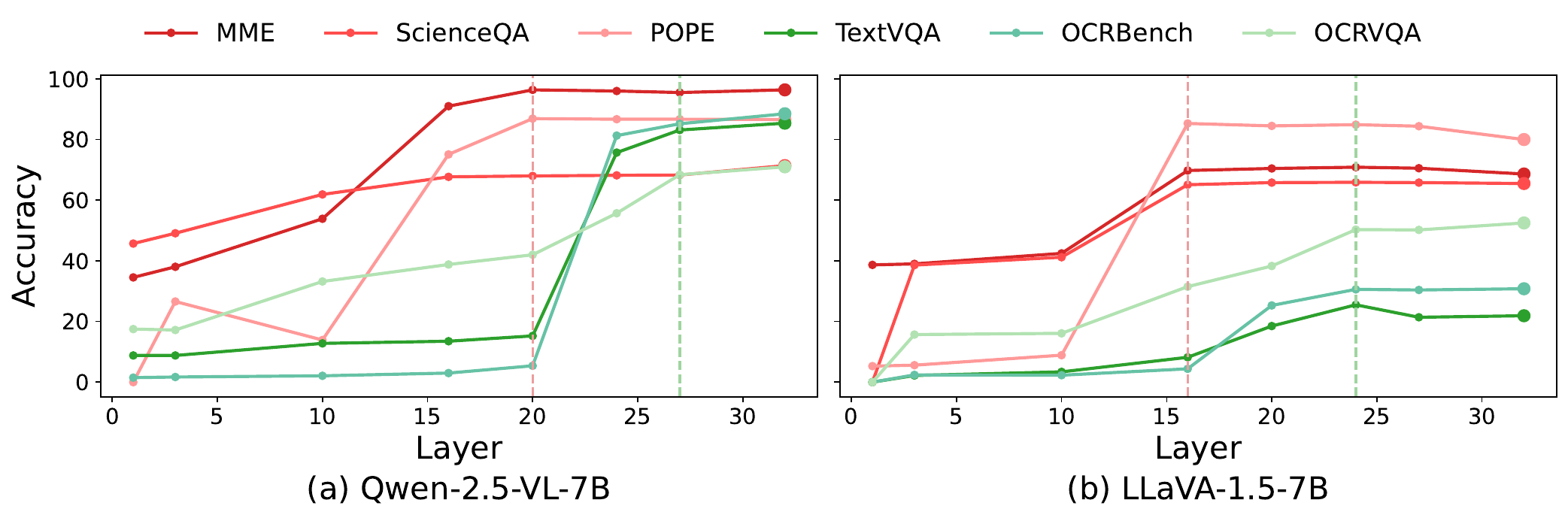}
    \vspace{-7mm}
    \caption{\textbf{Performance of models when pruning all visual tokens at different decoder layers}. Colored vertical dashed lines mark the convergence layers for datasets of different visual complexity, beyond which pruning all visual tokens has little impact on performance. This demonstrates that the layer at which visual tokens could be entirely removed depends on task visual complexity and model visual capability.
    }
    \vspace{-3mm}
    \label{fig:prune_all}
\end{figure}

\renewcommand{\multirowsetup}{\centering}
\begin{table*}
    \centering
    \vspace{-1mm}
    \setlength{\tabcolsep}{2.8pt}
    \renewcommand{\arraystretch}{1.33}
    \footnotesize
	\centering
	\caption{\textbf{Performance of Qwen2.5-VL-7B under varying pruning strategies}. Integrating random pruning with existing pruning methods leads to improved performance across multiple benchmarks. \textbf{Acc.} denotes the average accuracy across 7 benchmarks, \textbf{Rel.} represents the relative performance retained compared to the original model.}
    \vspace{-2mm}
	\label{tab:qwen}
    \begin{tabular}{p{2.8cm}|c c c c c c c | p{1.53cm} | p{1.85cm} }
        \shline
        \textbf{Method} & \textbf{MME} & \textbf{SQA} & \textbf{VQA}$^{\text{Text}}$  & \textbf{MMB} & \textbf{InfoVQA} & \textbf{DocVQA} & \textbf{OCRBench}   &\makecell[c]{\textbf{Acc. }}  &\makecell[c]{\textbf{Rel. (\%)}}  \\
        \shline
        \rowcolor{mygray}
        \multicolumn{10}{c}{\textit{Upper Bound} \ $\textbf{(100\%)}$}\\
        Qwen2.5-VL-7B & 2313 & 71.4 & 85.4  & 79.8 & 82.3 & 94.9 & 88.5   & \makecell[c]{83.6} & \makecell[c]{100.0}  \\ 

        \hline
        
        \rowcolor{mygray}
        \multicolumn{10}{c}{\textit{Retain 50\% Tokens} \ $\fg{(\downarrow 50\%)}$}\\
        \hline
        DART \texttt{\scriptsize{(EMNLP25)}} & 2295 & 69.2 & 82.1 & 79.6 & 64.4 & 88.5 & 75.5  & \makecell[c]{77.3} & \makecell[c]{92.7 $(\downarrow 7.3)$}  \\
        DivPrune \texttt{\scriptsize{(CVPR25)}} & 2291 & 70.2 & \textbf{83.1} & 79.4 & 73.9 & 92.7 & \textbf{84.1}  & \makecell[c]{80.7} & \makecell[c]{96.7 $(\downarrow 3.3)$}  \\
        \hline
        DART+VTW & 2302 & 69.8 & 76.3 & 79.4 & 65.8 & 86.5 & 73.3  & \makecell[c]{76.2} & \makecell[c]{91.4 $(\downarrow 8.6)$}  \\   
        DivPrune+VTW & 2308 & 70.0 & 75.8 & 79.4 & 73.0 & 89.8 & 79.9  & \makecell[c]{78.6} & \makecell[c]{94.2 $(\downarrow 5.8)$}  \\ 
        \hline
        DART+Random & \textbf{2318} & 70.0 & 82.7 & 79.6 & 65.9 & 89.3 & 77.9  & \makecell[c]{78.3} & \makecell[c]{93.9 $(\downarrow 6.1)$}  \\
        DivPrune+Random & 2302 & \textbf{70.3} & 82.8 & \textbf{79.9} & \textbf{74.9} & \textbf{92.9} & 83.3  & \makecell[c]{\textbf{80.9}} & \makecell[c]{\textbf{96.9} $(\downarrow 3.1)$}  \\
        \hline

        \shline
	\end{tabular}
     \vspace{-5mm}
\end{table*}

\subsection{Effectiveness of random pruning}
\label{sec:random_prune}
\paragraph{Combining with random pruning improves existing pruning methods.} 
As illustrated in \cref{fig:meaninfo_layerprune}, tokens in the first 10 layers contain more information than those in deeper layers. 
In this section, we demonstrate that applying existing pruning techniques to retain high-information tokens in shallow layers, while employing random pruning to eliminate less informative tokens in deeper layers, can better maintain model performance. 
% Specifically, we select three representative pruning techniques, including FastV~\cite{chen2024image}, DivPrune~\cite{alvar2025divprune}, and DART~\cite{wen2025stop}, and conduct comprehensive experiments on two VLLM architectures (Qwen-2.5-VL and LLaVA-1.5) across multiple benchmarks with varying visual complexity. More experiment details are provided in \cref{app:exp_details}.
Specifically, we select three representative pruning techniques, including FastV~\cite{chen2024image}, DivPrune~\cite{alvar2025divprune}, and DART~\cite{wen2025stop}. We also incorporate two multi-layer pruning methods, namely SparseVLM~\cite{zhang2024sparsevlm} and PDrop~\cite{xing2024pyramiddrop} for comparison. 
Furthermore, VTW~\cite{lin2025boosting} is included as it withdraws all visual tokens beyond a fixed layer, providing a contrast to our method that retains partial tokens in deeper layers.
Experiments are conducted on two VLLM architectures (Qwen-2.5-VL and LLaVA-1.5) across multiple benchmarks with varying visual complexity. More experiment details are provided in \cref{app:exp_details}.

As depicted in \cref{tab:qwen}, integrating random pruning enhances DART's performance on OCRBench from 75.5\% to 77.9\% given the same pruning ratio, maintaining 1.2\% more average original Qwen-2.5-VL-7B performance across 7 evaluated benchmarks (92.7\% \textit{vs.} 93.9\%).
Similarly, on LLaVA-1.5-7B, DivPrune + Random pruning achieves 61.3\% accuracy on MMBench when pruning 88.9\% visual tokens, which is a 6.7\% improvement over using only DivPrune (see \cref{tab:llava}).
Compared to multi-layer pruning methods like PDrop and SparseVLM, our methods are still competitive: at an 88.9\% pruning ratio, combining random pruning with FastV achieves 5.5\% higher relative performance over PDrop, while DART + Random pruning yields a 4.2\% relative performance improvement over SparseVLM.
% DART+random consistently outperforms PDrop and SparseVLM on LLaVA-1.5-7B under two pruning ratios, demonstrating the advantage of our hybrid strategy over existing multi-layer pruning methods.
\vspace{-1em}
\paragraph{Random pruning \textit{vs.}visual token withdraw.}
Furthermore, we compare the random pruning with VTW under the same pruning ratio.
According to \cref{tab:llava}, removing all visual tokens at a fixed layer results in suboptimal performance across 7 evaluated benchmark and 2 pruning ratios (DART + Random pruning achieves 98.2\% and 91.5\% relative performance compared to 83.2\% and 59.3\% for VTW). 
According to \cref{tab:qwen}, combining existing pruning methods with VTW, which prunes visual tokens in shallow layers using existing methods while removing all visual tokens in deeper layers, results in suboptimal performance on the TextVQA benchmark. Specifically, DART + Random pruning achieves 82.7\% accuracy, compared to 76.3\% for DART + VTW.
This trend is evident across other visually complex benchmarks, which depend on information from deeper layers.  
For example, DivPrune + Random pruning achieves 3.1\% and 3.4\% higher accuracy compared to the VTW counterpart on DocVQA and OCRBench, respectively.

\renewcommand{\multirowsetup}{\centering}
\begin{table}[t]
    \centering
    \setlength{\tabcolsep}{1pt}
    \renewcommand{\arraystretch}{1.33}
    \footnotesize
    \centering	
    % \vspace{-2mm}
    \caption{\textbf{Performance of LLaVA-1.5-7B under varying pruning strategies.}}
    \vspace{-2mm}
	\label{tab:llava}
    \resizebox{\columnwidth}{!}{
    \begin{tabular}{p{2.5cm}|c c c c c c c | c }
        \shline
        \textbf{Method} & \textbf{MME} & \textbf{SQA} & \textbf{POPE} & \textbf{VQA}$^{\text{Text}}$  & \textbf{MMB} & \textbf{GQA} & \textbf{MMB-CN}  &\makecell[c]{\textbf{Rel.}}  \\
        \shline
        \rowcolor{mygray}
        \multicolumn{9}{c}{\textit{Upper Bound, 576 Tokens} \ $\textbf{(100\%)}$}\\
        LLaVA-1.5-7B & 1862 & 69.5 & 85.9 & 58.3 & 64.6 & 62 & 58.1  & \makecell[c]{100.0}  \\ 

        \hline

        \rowcolor{mygray}
        \multicolumn{9}{c}{\textit{Retain 192 Tokens} \ $\fg{(\downarrow 66.7\%)}$} \\
        PDrop \texttt{\scriptsize{(CVPR25)}} & 1765 & 69.2 & 79.7 & 56.2 & 63.3 & 57.2 & 56.5  & \makecell[c]{95.9 }  \\
        \scriptsize{SparseVLM} \texttt{\scriptsize{(ICML25)}}& 1779 & 68.6 & 84.9 & \textbf{57.7} & 63.4 & 58.9 & 57.1  & \makecell[c]{97.6}  \\
        VTW \texttt{\scriptsize{(AAAI25)}}& 1533 & 67.6 & 51.3 & 48.1 & 61.2 & 46.3 & 53  & \makecell[c]{83.2 }  \\
        FastV \texttt{\scriptsize{(ECCV24)}} & 1796 & 69.1 & 78.3 & 56.4 & 63.2 & \textbf{57.7} & 57.4  & \makecell[c]{96.2 }  \\
        DART \texttt{\scriptsize{(EMNLP25)}}& 1833 & 69.0 & 81.7 & 57.0 & 63.7 & 59.5 & 57.1  & \makecell[c]{97.6 } \\
        DivPrune \texttt{\scriptsize{(CVPR25)}} & 1769 & 68.6 & \textbf{87.6} & 56.6 & 62.3 & \textbf{60.1} & 56.2  & \makecell[c]{97.6 }  \\
        \hline
        FastV+VTW & 1848 & 68.6 & 81.8 & 53.6 & \textbf{64.7} & 57.5 & 57.6  & \makecell[c]{96.7}  \\
        DART+VTW & 1807 & 69.1 & 84.4 & 56.4 & 64.1 & 59.1 & \textbf{57.7}  & \makecell[c]{97.9 }  \\
        DivPrune+VTW & 1772 & 69.1 & 87.0 & 56.0 & 63.2 & 60.0 & 57.2  & \makecell[c]{97.9 }  \\
        \hline
        FastV+Random & \textbf{1853} & 68.6 & 81.5 & 53.5 & \textbf{64.7} & 57.8 & 57.5  & \makecell[c]{96.7}  \\
        DART+Random & 1816 & 69.2 & 84.4 & 56.8 & 63.9 & 59.6 & 57.7  & \makecell[c]{\textbf{98.2} }  \\
        DivPrune+Random & 1777 & \textbf{69.3} & 87.1 & 55.9 & 63.2 & \textbf{60.1} & 56.9  & \makecell[c]{97.9 } \\
        
        \hline

        \rowcolor{mygray}
        \multicolumn{9}{c}{\textit{Retain 64 Tokens} \ $\fg{(\downarrow 88.9\%)}$}\\
        PDrop \texttt{\scriptsize{(CVPR25)}} & 1263 & 68.0 & 45.7 & 48.3 & 45.9 & 45.5 & 34.9 & \makecell[c]{72.3} \\
        \scriptsize{SparseVLM} \texttt{\scriptsize{(ICML25)}}& 1487 & 68.7 & 69.9 & 52.6 & 58.6 & 51.9 & 50.3 & \makecell[c]{87.3} \\
        VTW \texttt{\scriptsize{(AAAI25)}}& 953 & 65.2 & 25.2 & 43 & 22 & 38.9 & 40.6  & \makecell[c]{59.3}  \\
        FastV \texttt{\scriptsize{(ECCV24)}} & 1088 & 68.2 & 23.0 & 46.0 & 34.1 & 41.4 & 25.9& \makecell[c]{60.9} \\ 
        DART \texttt{\scriptsize{(EMNLP25)}}& 1529 & 68.6 & 59.6 & 50.4 & 55.8 & 51.2 & 46.7 & \makecell[c]{83.7 } \\
        DivPrune \texttt{\scriptsize{(CVPR25)}} & 1614 & 67.9 & 85.6 & \textbf{55.0} & 54.6 & 57.7 & 52.8 & \makecell[c]{92.4} \\
        \hline
        FastV+VTW & 1361 & 70.4 & 42.1 & 47.0 & 54.6 & 47.1 & 46.4 & \makecell[c]{77.8} \\
        DART+VTW & 1664 & 68.3 & 74.0 & 53.3 & 60.3 & 55.5 & 53.1 & \makecell[c]{91.3 } \\
        DivPrune+VTW & 1677 & 68.2 & \textbf{86.8} & 54.1 & \textbf{61.6} & 57.1 & 53.7 & \makecell[c]{94.6} \\
        \hline
        FastV+Random & 1350 & \textbf{70.6} & 42.1 & 47.4 & 54.9 & 46.6 & 46.4 & \makecell[c]{77.8 } \\
        DART+Random & 1670 & 68.4 & 74.1 & 53.4 & 60.4 & 55.1 & 53.4 & \makecell[c]{91.5 } \\
        DivPrune+Random & \textbf{1693} & 68.3 & \textbf{86.8} & 54.5 & 61.3 & \textbf{57.8} & \textbf{53.8} & \makecell[c]{\textbf{94.9}} \\
        \shline
	\end{tabular}
    }
     \vspace{-7mm}
\end{table}

\subsection{Efficiency analysis}
\label{sec:eff}

To demonstrate the efficiency of integrating existing pruning methods with random pruning, we perform a comparative analysis of FLOPs, cache storage, CUDA time and inference latency on LLaVA-1.5-7B.  All experiments are performed on a single NVIDIA A6000-40GB GPU, evaluated using the TextVQA benchmark. As shown in \cref{tab:efficiency}, by retaining only 64 tokens, combining DART with Random pruning reduces the inference latency to $\mathbf{0.6\times}$ the original, while retaining 91.6\% of the model performance. Compared to FastV with the same reduction ratio, DART+Random pruning requires less memory, achieves faster inference and better performance. Diversity-based pruning methods enable compatibility with faster attention implementations like FlashAttention, which are infeasible for methods like FastV that require access to the text-visual attention information with in the language model. Moreover, integrating random pruning reduces DART's FLOPs from 2.44T to 2.36T while enhances its performance from 50.4\% to 53.4\%, under an average retained token number of 64. When compared to removing all visual tokens at a specified layer, DART+Random pruning yields both faster inference and higher performance. For example, DART+Random pruning achieves 56.8\% accuracy and requires 4.06T FLOPs compared to 56.4\% and 4.08T of the VTW counterpart, under the same pruning ratio of 192 tokens.

\subsection{Comparison with Prior Observations}
\label{sec:compare_prior}

Previous studies, such as VTW and PDrop, have highlighted the redundancy of visual tokens in the deep layers of VLMs.
VTW observes that the attention values from the last text token to visual tokens approach zero in deep layers, and therefore proposes to withdraw all visual tokens beyond a fixed layer across different models and datasets.
PDrop systematically examines the impact of pruning visual tokens at different layers and finds that model performance degrades less when pruning occurs in deep layers, suggesting that token redundancy increases with depth. Based on this observation, it introduces a layer-wise pruning method with higher pruning ratios in deeper layers, guided by attention values similar to FastV.

While these works reveal a common phenomenon of token redundancy, our study revisits this problem from an information perspective and arrives at a distinct conclusion. We find that the depth at which visual tokens lose their information, termed as ``information horizon", is not fixed but rather correlates with both the model capacity and the visual complexity of the task. Consequently, unlike VTW which removes all visual tokens beyond a predetermined layer, we propose to retain a small subset of tokens in the deep layers, achieving a better trade-off between efficiency and performance. Moreover, our quantitative analysis reveals that in deep layers, the information of all visual tokens converges toward zero. This explains why attention-based methods perform no better than random pruning at these layers. Guided by this insight, we adopt random pruning in the deep layer, which preserves accuracy while removing the need for any computation and can be seamlessly integrated with FlashAttention for faster inference.

\begin{table}[t]
  \centering
  \caption{\textbf{Efficiency analysis with LLaVA-1.5-7B on TextVQA.} At the same reduction ratio, DART+Random pruning requires less memory and achieves faster inference compared to other methods.}
  \vspace{-2mm}
  \resizebox{\linewidth}{!}{
    \begin{tabular}{l|c|c|c|c|c|c}
    \toprule
    \textbf{Method}        & \textbf{\# Token}              & \begin{tabular}[c]{@{}c@{}}\textbf{FLOPs}\\ \textbf{(T)}\end{tabular} & \begin{tabular}[c]{@{}c@{}}\textbf{Storage}\\ \textbf{(MB)}\end{tabular} & \begin{tabular}[c]{@{}c@{}}\textbf{CUDA Time}\\ \textbf{(ms)}\end{tabular} & \begin{tabular}[c]{@{}c@{}}\textbf{Latency}\\ \textbf{(ms)}\end{tabular} & \begin{tabular}[c]{@{}c@{}}\textbf{Acc.}\\ \textbf{(\%)}\end{tabular}\\
    \midrule
    LLaVA-1.5-7B & 576 & 9.22 & 346.2  & 127.4 & 272.4 & 58.3\\
    \midrule
    FastV{\small\texttt{(ECCV24)}} & \multirow{4}{*}{192} & 4.18 & 159.1  & 76.7 & 224.2 & 56.4\\
    DART{\small\texttt{(EMNLP25)}}    &                      & 4.12 & 156.8  & 75 & 218.2 & \textbf{57.0} \\
    DART+VTW    &                      & 4.08 & \textbf{154.0}  & 67.6 & 184.6 & 56.4\\
    DART+Random    &                      & \textbf{4.06} & 154.2  & \textbf{67.4} & \textbf{183.6} & 56.8\\
    \midrule
    FastV{\small\texttt{(ECCV24)}} & \multirow{4}{*}{64} & 2.59 & 98.9 & 55.7 & 202.0 & 46.0\\
    DART{\small\texttt{(EMNLP25)}}    &                      & 2.44 & 93.1 & 54.3 & 200.6 & 50.4\\
    DART+VTW    &                      & 2.37 & 90.7 & 47.5 & 164.2 & 53.3 \\
    DART+Random    &                      & \textbf{2.36} & \textbf{90.1} & \textbf{47.1} & \textbf{162.2} & \textbf{53.4} \\
    \bottomrule
    \end{tabular}
  }
  \vspace{-7mm}
  \label{tab:efficiency}
\end{table}
\section{Conclusion}

In this paper, we investigate the visual token information in VLLMs, which is defined as the change of output probability when a visual token is removed.
Our experimental findings reveal that visual token information diminishes to almost zero at a particular layer, influenced by task visual complexity and model visual capability.
This explains why current token pruning methods do not surpass random pruning at deep layers, showing that integrating random pruning yields superior results across various models and benchmarks.
We believe our findings offer empirical insights for employing informative visual tokens to enhance VLLMs.

\section{Acknowledgments}

This work was supported by the Yeqisun Joint Funds of the National Natural Science Foundation of China under Grant U2441252, the National Natural Science Foundation of China under Grant 62571372, 62331001, 62271155, the Shanghai Municipal Science and Technology Major Project 2021SHZDZX0100, the Changjiang Scholars Program of China, the Fundamental Research Funds for the Central Universities, the Shanghai Key Technology R\&D Plan-``Computational Biology" Project 25JS2840100.
{
    \small
    \bibliographystyle{ieeenat_fullname}
    \bibliography{main}

@String(CVPR= {IEEE Conf. Comput. Vis. Pattern Recog.})

@String(ICLR = {Int. Conf. Learn. Represent.})

@String(AAAI = {AAAI})

@String(CVPR  = {CVPR})

@String(ICLR  = {ICLR})

@article{bai2025qwen2,
  title={Qwen2. 5-vl technical report},
  author={Bai, Shuai and Chen, Keqin and Liu, Xuejing and Wang, Jialin and Ge, Wenbin and Song, Sibo and Dang, Kai and Wang, Peng and Wang, Shijie and Tang, Jun and others},
  journal={arXiv preprint arXiv:2502.13923},
  year={2025}
}

@article{liu2023visual,
  title={Visual instruction tuning},
  author={Liu, Haotian and Li, Chunyuan and Wu, Qingyang and Lee, Yong Jae},
  journal={Advances in neural information processing systems},
  volume={36},
  pages={34892--34916},
  year={2023}
}

@article{li2024llava,
  title={Llava-onevision: Easy visual task transfer},
  author={Li, Bo and Zhang, Yuanhan and Guo, Dong and Zhang, Renrui and Li, Feng and Zhang, Hao and Zhang, Kaichen and Zhang, Peiyuan and Li, Yanwei and Liu, Ziwei and others},
  journal={arXiv preprint arXiv:2408.03326},
  year={2024}
}

@article{chen2025sft,
  title={Sft or rl? an early investigation into training r1-like reasoning large vision-language models},
  author={Chen, Hardy and Tu, Haoqin and Wang, Fali and Liu, Hui and Tang, Xianfeng and Du, Xinya and Zhou, Yuyin and Xie, Cihang},
  journal={arXiv preprint arXiv:2504.11468},
  year={2025}
}

@article{lu2022learn,
  title={Learn to explain: Multimodal reasoning via thought chains for science question answering},
  author={Lu, Pan and Mishra, Swaroop and Xia, Tanglin and Qiu, Liang and Chang, Kai-Wei and Zhu, Song-Chun and Tafjord, Oyvind and Clark, Peter and Kalyan, Ashwin},
  journal={Advances in Neural Information Processing Systems},
  volume={35},
  pages={2507--2521},
  year={2022}
}

@misc{fu2024mmecomprehensiveevaluationbenchmark,
      title={MME: A Comprehensive Evaluation Benchmark for Multimodal Large Language Models}, 
      author={Chaoyou Fu and Peixian Chen and Yunhang Shen and Yulei Qin and Mengdan Zhang and Xu Lin and Jinrui Yang and Xiawu Zheng and Ke Li and Xing Sun and Yunsheng Wu and Rongrong Ji},
      year={2024},
      eprint={2306.13394},
      archivePrefix={arXiv},
      primaryClass={cs.CV},
      url={https://arxiv.org/abs/2306.13394}, 
}

@inproceedings{singh2019towards,
  title={Towards vqa models that can read},
  author={Singh, Amanpreet and Natarajan, Vivek and Shah, Meet and Jiang, Yu and Chen, Xinlei and Batra, Dhruv and Parikh, Devi and Rohrbach, Marcus},
  booktitle={Proceedings of the IEEE/CVF conference on computer vision and pattern recognition},
  pages={8317--8326},
  year={2019}
}

@article{rajabi2024gsr,
  title={Gsr-bench: A benchmark for grounded spatial reasoning evaluation via multimodal llms},
  author={Rajabi, Navid and Kosecka, Jana},
  journal={arXiv preprint arXiv:2406.13246},
  year={2024}
}

@article{zhang2024mme,
  title={Mme-realworld: Could your multimodal llm challenge high-resolution real-world scenarios that are difficult for humans?},
  author={Zhang, Yi-Fan and Zhang, Huanyu and Tian, Haochen and Fu, Chaoyou and Zhang, Shuangqing and Wu, Junfei and Li, Feng and Wang, Kun and Wen, Qingsong and Zhang, Zhang and others},
  journal={arXiv preprint arXiv:2408.13257},
  year={2024}
}

@inproceedings{chen2024image,
  title={An image is worth 1/2 tokens after layer 2: Plug-and-play inference acceleration for large vision-language models},
  author={Chen, Liang and Zhao, Haozhe and Liu, Tianyu and Bai, Shuai and Lin, Junyang and Zhou, Chang and Chang, Baobao},
  booktitle={European Conference on Computer Vision},
  pages={19--35},
  year={2024},
  organization={Springer}
}

@inproceedings{zhang2024sparsevlm,
  title={SparseVLM: Visual Token Sparsification for Efficient Vision-Language Model Inference},
  author={Zhang, Yuan and Fan, Chun-Kai and Ma, Junpeng and Zheng, Wenzhao and Huang, Tao and Cheng, Kuan and Gudovskiy, Denis and Okuno, Tomoyuki and Nakata, Yohei and Keutzer, Kurt and others},
  booktitle={International Conference on Machine Learning},
  year={2025}
}

@inproceedings{ye2025fit,
  title={Fit and prune: Fast and training-free visual token pruning for multi-modal large language models},
  author={Ye, Weihao and Wu, Qiong and Lin, Wenhao and Zhou, Yiyi},
  booktitle={Proceedings of the AAAI Conference on Artificial Intelligence},
  volume={39},
  pages={22128--22136},
  year={2025}
}

@article{liu2024multi,
  title={Multi-stage vision token dropping: Towards efficient multimodal large language model},
  author={Liu, Ting and Shi, Liangtao and Hong, Richang and Hu, Yue and Yin, Quanjun and Zhang, Linfeng},
  journal={arXiv preprint arXiv:2411.10803},
  year={2024}
}

@inproceedings{alvar2025divprune,
  title={Divprune: Diversity-based visual token pruning for large multimodal models},
  author={Alvar, Saeed Ranjbar and Singh, Gursimran and Akbari, Mohammad and Zhang, Yong},
  booktitle={Proceedings of the Computer Vision and Pattern Recognition Conference},
  pages={9392--9401},
  year={2025}
}

@article{wen2025stop,
  title={Stop looking for important tokens in multimodal language models: Duplication matters more},
  author={Wen, Zichen and Gao, Yifeng and Wang, Shaobo and Zhang, Junyuan and Zhang, Qintong and Li, Weijia and He, Conghui and Zhang, Linfeng},
  journal={arXiv preprint arXiv:2502.11494},
  year={2025}
}

@inproceedings{liu2024improved,
  title={Improved baselines with visual instruction tuning},
  author={Liu, Haotian and Li, Chunyuan and Li, Yuheng and Lee, Yong Jae},
  booktitle={Proceedings of the IEEE/CVF conference on computer vision and pattern recognition},
  pages={26296--26306},
  year={2024}
}

@inproceedings{zhao2025stitch,
  title={A stitch in time saves nine: Small vlm is a precise guidance for accelerating large vlms},
  author={Zhao, Wangbo and Han, Yizeng and Tang, Jiasheng and Li, Zhikai and Song, Yibing and Wang, Kai and Wang, Zhangyang and You, Yang},
  booktitle={Proceedings of the Computer Vision and Pattern Recognition Conference},
  pages={19814--19824},
  year={2025}
}

@inproceedings{dao2022flashattention,
  title={Flash{A}ttention: Fast and Memory-Efficient Exact Attention with {IO}-Awareness},
  author={Dao, Tri and Fu, Daniel Y. and Ermon, Stefano and Rudra, Atri and R{\'e}, Christopher},
  booktitle={Advances in Neural Information Processing Systems (NeurIPS)},
  year={2022}
}

@inproceedings{dao2023flashattention2,
  title={Flash{A}ttention-2: Faster Attention with Better Parallelism and Work Partitioning},
  author={Dao, Tri},
  booktitle={International Conference on Learning Representations (ICLR)},
  year={2024}
}

@article{liu2024ocrbench,
  title={Ocrbench: on the hidden mystery of ocr in large multimodal models},
  author={Liu, Yuliang and Li, Zhang and Huang, Mingxin and Yang, Biao and Yu, Wenwen and Li, Chunyuan and Yin, Xu-Cheng and Liu, Cheng-Lin and Jin, Lianwen and Bai, Xiang},
  journal={Science China Information Sciences},
  volume={67},
  number={12},
  pages={220102},
  year={2024},
  publisher={Springer}
}

@inproceedings{mishra2019ocr,
  title={Ocr-vqa: Visual question answering by reading text in images},
  author={Mishra, Anand and Shekhar, Shashank and Singh, Ajeet Kumar and Chakraborty, Anirban},
  booktitle={2019 international conference on document analysis and recognition (ICDAR)},
  pages={947--952},
  year={2019},
  organization={IEEE}
}

@inproceedings{li2023evaluating,
  title={Evaluating Object Hallucination in Large Vision-Language Models},
  author={Yifan Li and Yifan Du and Kun Zhou and Jinpeng Wang and Wayne Xin Zhao and Ji-Rong Wen},
  booktitle={The 2023 Conference on Empirical Methods in Natural Language Processing},
  year={2023},
  url={https://openreview.net/forum?id=xozJw0kZXF}
}

@inproceedings{zhai2023sigmoid,
  title={Sigmoid loss for language image pre-training},
  author={Zhai, Xiaohua and Mustafa, Basil and Kolesnikov, Alexander and Beyer, Lucas},
  booktitle={Proceedings of the IEEE/CVF international conference on computer vision},
  pages={11975--11986},
  year={2023}
}

@inproceedings{radford2021learning,
  title={Learning transferable visual models from natural language supervision},
  author={Radford, Alec and Kim, Jong Wook and Hallacy, Chris and Ramesh, Aditya and Goh, Gabriel and Agarwal, Sandhini and Sastry, Girish and Askell, Amanda and Mishkin, Pamela and Clark, Jack and others},
  booktitle={International conference on machine learning},
  pages={8748--8763},
  year={2021},
  organization={PmLR}
}

@inproceedings{lin2025boosting,
  title={Boosting multimodal large language models with visual tokens withdrawal for rapid inference},
  author={Lin, Zhihang and Lin, Mingbao and Lin, Luxi and Ji, Rongrong},
  booktitle={Proceedings of the AAAI Conference on Artificial Intelligence},
  volume={39},
  pages={5334--5342},
  year={2025}
}

@inproceedings{liu2024mmbench,
  title={Mmbench: Is your multi-modal model an all-around player?},
  author={Liu, Yuan and Duan, Haodong and Zhang, Yuanhan and Li, Bo and Zhang, Songyang and Zhao, Wangbo and Yuan, Yike and Wang, Jiaqi and He, Conghui and Liu, Ziwei and others},
  booktitle={European conference on computer vision},
  pages={216--233},
  year={2024},
  organization={Springer}
}

@article{hudson2018gqa,
    title={GQA: A New Dataset for Real-World Visual Reasoning 
    and Compositional Question Answering},
    author={Hudson, Drew A and Manning, Christopher D},
    journal={Conference on Computer Vision and Pattern Recognition (CVPR)},
    year={2019}
}

@inproceedings{mathew2022infographicvqa,
  title={Infographicvqa},
  author={Mathew, Minesh and Bagal, Viraj and Tito, Rub{\`e}n and Karatzas, Dimosthenis and Valveny, Ernest and Jawahar, CV},
  booktitle={Proceedings of the IEEE/CVF Winter Conference on Applications of Computer Vision},
  pages={1697--1706},
  year={2022}
}

@inproceedings{mathew2021docvqa,
  title={Docvqa: A dataset for vqa on document images},
  author={Mathew, Minesh and Karatzas, Dimosthenis and Jawahar, CV},
  booktitle={Proceedings of the IEEE/CVF winter conference on applications of computer vision},
  pages={2200--2209},
  year={2021}
}

@inproceedings{bolya2022tome,
  title={Token Merging: Your {ViT} but Faster},
  author={Bolya, Daniel and Fu, Cheng-Yang and Dai, Xiaoliang and Zhang, Peizhao and Feichtenhofer, Christoph and Hoffman, Judy},
  booktitle={International Conference on Learning Representations},
  year={2023}
}

@article{wang2025folder,
  title={FOLDER: Accelerating Multi-modal Large Language Models with Enhanced Performance},
  author={Wang, Haicheng and Yu, Zhemeng and Spadaro, Gabriele and Ju, Chen and Qu{\'e}tu, Victor and Tartaglione, Enzo},
  journal={arXiv preprint arXiv:2501.02430},
  year={2025}
}

@article{zhu2023minigpt,
  title={MiniGPT-4: Enhancing Vision-Language Understanding with Advanced Large Language Models},
  author={Zhu, Deyao and Chen, Jun and Shen, Xiaoqian and Li, Xiang and Elhoseiny, Mohamed},
  journal={arXiv preprint arXiv:2304.10592},
  year={2023}
}

@inproceedings{cheng2024spatialrgpt,
  title={SpatialRGPT: Grounded Spatial Reasoning in Vision-Language Models},
  author={Cheng, An-Chieh and Yin, Hongxu and Fu, Yang and Guo, Qiushan and Yang, Ruihan and Kautz, Jan and Wang, Xiaolong and Liu, Sifei},
  booktitle={NeurIPS},
  year={2024}
}

@inproceedings{ju2024turbo,
  title={Turbo: Informativity-driven acceleration plug-in for vision-language large models},
  author={Ju, Chen and Wang, Haicheng and Cheng, Haozhe and Chen, Xu and Zhai, Zhonghua and Huang, Weilin and Lan, Jinsong and Xiao, Shuai and Zheng, Bo},
  booktitle={European Conference on Computer Vision},
  pages={436--455},
  year={2024},
  organization={Springer}
}

@article{zhang2025cdpruner,
  title={Beyond Attention or Similarity: Maximizing Conditional Diversity for Token Pruning in MLLMs},
  author={Zhang, Qizhe and Liu, Mengzhen and Li, Lichen and Lu, Ming and Zhang, Yuan and Pan, Junwen and She, Qi and Zhang, Shanghang},
  journal={arXiv preprint arXiv:2506.10967},
  year={2025}
}

@misc{li2025todrevisualtokenpruning,
      title={ToDRE: Visual Token Pruning via Diversity and Task Awareness for Efficient Large Vision-Language Models}, 
      author={Duo Li and Zuhao Yang and Shijian Lu},
      year={2025},
      eprint={2505.18757},
      archivePrefix={arXiv},
      primaryClass={cs.CV},
      url={https://arxiv.org/abs/2505.18757}, 
}

@inproceedings{ma2024groma,
  title={Groma: Localized visual tokenization for grounding multimodal large language models},
  author={Ma, Chuofan and Jiang, Yi and Wu, Jiannan and Yuan, Zehuan and Qi, Xiaojuan},
  booktitle={European Conference on Computer Vision},
  pages={417--435},
  year={2024},
  organization={Springer}
}

@article{fu2025livevqa,
  title={LiveVQA: Live Visual Knowledge Seeking},
  author={Fu, Mingyang and Peng, Yuyang and Liu, Benlin and Wan, Yao and Chen, Dongping},
  journal={arXiv preprint arXiv:2504.05288},
  year={2025}
}

@article{xing2024pyramiddrop,
  title={Pyramiddrop: Accelerating your large vision-language models via pyramid visual redundancy reduction},
  author={Xing, Long and Huang, Qidong and Dong, Xiaoyi and Lu, Jiajie and Zhang, Pan and Zang, Yuhang and Cao, Yuhang and He, Conghui and Wang, Jiaqi and Wu, Feng and others},
  journal={arXiv preprint arXiv:2410.17247},
  year={2024}
}

@article{huang2025medvlthinker,
  title={MedVLThinker: Simple Baselines for Multimodal Medical Reasoning},
  author={Huang, Xiaoke and Wu, Juncheng and Liu, Hui and Tang, Xianfeng and Zhou, Yuyin},
  journal={arXiv preprint arXiv:2508.02669},
  year={2025}
}

@inproceedings{chen2024internvl,
  title={Internvl: Scaling up vision foundation models and aligning for generic visual-linguistic tasks},
  author={Chen, Zhe and Wu, Jiannan and Wang, Wenhai and Su, Weijie and Chen, Guo and Xing, Sen and Zhong, Muyan and Zhang, Qinglong and Zhu, Xizhou and Lu, Lewei and others},
  booktitle={Proceedings of the IEEE/CVF Conference on Computer Vision and Pattern Recognition},
  pages={24185--24198},
  year={2024}
}

@misc{vicuna2023,
    title = {Vicuna: An Open-Source Chatbot Impressing GPT-4 with 90\%* ChatGPT Quality},
    url = {https://lmsys.org/blog/2023-03-30-vicuna/},
    author = {Chiang, Wei-Lin and Li, Zhuohan and Lin, Zi and Sheng, Ying and Wu, Zhanghao and Zhang, Hao and Zheng, Lianmin and Zhuang, Siyuan and Zhuang, Yonghao and Gonzalez, Joseph E. and Stoica, Ion and Xing, Eric P.},
    month = {March},
    year = {2023}
}

@misc{touvron2023llamaopenefficientfoundation,
      title={LLaMA: Open and Efficient Foundation Language Models}, 
      author={Hugo Touvron and Thibaut Lavril and Gautier Izacard and Xavier Martinet and Marie-Anne Lachaux and Timothée Lacroix and Baptiste Rozière and Naman Goyal and Eric Hambro and Faisal Azhar and Aurelien Rodriguez and Armand Joulin and Edouard Grave and Guillaume Lample},
      year={2023},
      eprint={2302.13971},
      archivePrefix={arXiv},
      primaryClass={cs.CL},
      url={https://arxiv.org/abs/2302.13971}, 
}

@misc{bai2023qwentechnicalreport,
      title={Qwen Technical Report}, 
      author={Jinze Bai and Shuai Bai and Yunfei Chu and Zeyu Cui and Kai Dang and Xiaodong Deng and Yang Fan and Wenbin Ge and Yu Han and Fei Huang and Binyuan Hui and Luo Ji and Mei Li and Junyang Lin and Runji Lin and Dayiheng Liu and Gao Liu and Chengqiang Lu and Keming Lu and Jianxin Ma and Rui Men and Xingzhang Ren and Xuancheng Ren and Chuanqi Tan and Sinan Tan and Jianhong Tu and Peng Wang and Shijie Wang and Wei Wang and Shengguang Wu and Benfeng Xu and Jin Xu and An Yang and Hao Yang and Jian Yang and Shusheng Yang and Yang Yao and Bowen Yu and Hongyi Yuan and Zheng Yuan and Jianwei Zhang and Xingxuan Zhang and Yichang Zhang and Zhenru Zhang and Chang Zhou and Jingren Zhou and Xiaohuan Zhou and Tianhang Zhu},
      year={2023},
      eprint={2309.16609},
      archivePrefix={arXiv},
      primaryClass={cs.CL},
      url={https://arxiv.org/abs/2309.16609}, 
}

@misc{2023internlm,
    title={InternLM: A Multilingual Language Model with Progressively Enhanced Capabilities},
    author={InternLM Team},
    howpublished = {\url{https://github.com/InternLM/InternLM-techreport}},
    year={2023}
}

@article{wang2025internvl3_5,
  title={InternVL3.5: Advancing Open-Source Multimodal Models in Versatility, Reasoning, and Efficiency},
  author={Wang, Weiyun and Gao, Zhangwei and Gu, Lixin and Pu, Hengjun and Cui, Long and Wei, Xingguang and Liu, Zhaoyang and Jing, Linglin and Ye, Shenglong and Shao, Jie and others},
  journal={arXiv preprint arXiv:2508.18265},
  year={2025}
}
}

\clearpage
\setcounter{page}{1}
\maketitlesupplementary

% \section{Alternative Strategies for Computing Token Importance}
% \label{app:cal_tkn_imp}

\section{Experiment details}
\label{app:exp_details}
\textbf{Datasets.} We conduct experiments on nine widely adopted benchmarks, which we categorize into three groups to reflect varying levels of visual complexity: 
\emph{(i) knowledge QA}, including GQA~\cite{hudson2018gqa}, SQA (ScienceQA)~\cite{lu2022learn}, and POPE~\cite{li2023evaluating}, which require factual reasoning grounded in visual and textual information; 
\emph{(ii) text-centric understanding }, including VQA$^{\text{Text}}$ (TextVQA)~\cite{singh2019towards}, InfoVQA~\cite{mathew2022infographicvqa}, DocVQA~\cite{mathew2021docvqa}, and OCRBench~\cite{liu2024ocrbench}, which involve extracting and reasoning over fine-grained text in images, leading to higher visual complexity; 
\emph{(iii) Comprehensive evaluation}, including MME~\cite{fu2024mmecomprehensiveevaluationbenchmark}, MMB (MMBench) and MMB-CN (MMBench-CN)~\cite{liu2024mmbench}, which cover a broad range of multimodal tasks and provide a comprehensive assessment of a model's multimodal understanding capability. 

\textbf{VLLM Architectures.} We verify different pruning strategies on two representative VLLM architectures: 
LLaVA-1.5~\cite{liu2024improved}, which decodes an image into a fixed number of visual tokens, 
and Qwen2.5-VL~\cite{bai2025qwen2}, which adaptively determines the number of visual tokens based on the resolution of the input image.

\textbf{Combining existing methods with random pruning.} 
DivPrune removes visual tokens before they enter the language decoder, while DART and FastV perform pruning at the second and third layers, respectively. 
We study two hybrid pruning strategies: first, we apply $S \in \{\text{FastV}, \text{DART}, \text{DivPrune}\}$ at a shallow layer to prune part of the visual tokens. Subsequently, at a specified deep layer, an additional pruning step is performed: (i) $S$+VTW: following VTW~\cite{lin2025boosting}, all remaining visual tokens are withdrawn, or (ii) $S$+Random: random pruning is applied to further remove visual tokens.
Detailed pruning layers and ratios are illustrated in \cref{tab:setting_qwen} and \cref{tab:setting_llava}.
For pruning on Qwen2.5-VL-7B (28 decoder layers, with an average per-layer retention of 50\% visual tokens) and LLaVA-1.5-7B (32 decoder layers, with an average per-layer retention of 192 visual tokens), we follow the original methods for pruning at their respective shallow layers, while applying an additional pruning step at deeper layers: $S$+VTW prunes all tokens at the 26th and 21th layers for Qwen and LLaVA, respectively, whereas $S$+Random applies random pruning at the 25th and 20th layers.
For pruning on LLaVA-1.5-7B (with an average per-layer retention of 64 visual tokens), pruning at the second or third layer would retain too few tokens under this setting. For instance, DART need to prune 89\% of visual tokens while FastV need to  prune 98\%. Therefore, for the hybrid strategies based on DART and FastV, the shallow pruning layer is shifted one layer earlier to prevent pruning away an excessive number of tokens at the shallow pruning stage, while the shallow pruning layer remains is unchanged for hybrid strategies based on DivPrune.

\begin{table}[ht]
\centering
\caption{Pruning settings for Qwen2.5-VL-7B (28 decoder layers) with an average per-layer retention of 50\% visual tokens. Layer~0 indicates pruning before the visual tokens enter the language decoder.}
\label{tab:setting_qwen}
\begin{tabular}{l|c|c}
\shline
\rowcolor{mygray}
\multicolumn{3}{c}{\textit{Retain 50\% Tokens} \ $\fg{(\downarrow 50\%)}$}\\
Method & Layer & Retain ratio \\
\hline
DART                 & 2  & 0.46 \\
\multirow{2}{*}{DART+VTW} & 2 & 0.49    \\
                        & 26   &  0     \\
\multirow{2}{*}{DART+Random} & 2  & 0.49 \\
                        & 25 & 0.25 \\
\hline
DivPrune             & 0  & 0.50 \\
\multirow{2}{*}{DivPrune+VTW} & 0 & 0.53 \\
                        & 26  & 0  \\
\multirow{2}{*}{DivPrune+Random} & 0  & 0.53 \\
                        & 25 & 0.25 \\
\shline
\end{tabular}
\end{table}

\begin{table*}[t]
\centering
\caption{Pruning settings for LLaVA-1.5-7B (32 decoder layers) with an average per-layer retention of 192 and 64 visual tokens.}
\label{tab:setting_llava}
\begin{tabular}{l | c | c | c | c}
\shline
\rowcolor{mygray}
\multicolumn{1}{c}{ } 
& \multicolumn{2}{c}{\textit{Retain 192 Tokens} \ $\fg{(\downarrow 66.7\%)}$} 
& \multicolumn{2}{c}{\textit{Retain 64 Tokens} \ $\fg{(\downarrow 88.9\%)}$} \\
Method & Layer & Retain ratio & Layer & Retain ratio \\
\hline
DART                     & 2  & 0.29 & 2  & 0.05 \\
\multirow{2}{*}{DART+VTW}                & 2  & 0.44 & 1  & 0.10 \\
                         & 21 & 0    & 26 & 0    \\
\multirow{2}{*}{DART+Random}                   & 2  & 0.44 & 1  & 0.10 \\
                         & 20 & 0.07 & 20 & 0.05 \\
\hline
DivPrune                 & 0  & 0.33 & 0  & 0.11 \\
\multirow{2}{*}{DivPrune  +VTW}             & 0  & 0.49 & 0  & 0.17 \\
                         & 21 & 0    & 21 & 0    \\
\multirow{2}{*}{DivPrune  +Random}              & 0  & 0.49 & 0  & 0.17 \\
                         & 20 & 0.07 & 20 & 0.02 \\
\hline
FastV                    & 3  & 0.26 & 3  & 0.02 \\
\multirow{2}{*}{FastV+VTW}               & 3  & 0.41 & 2  & 0.06 \\
                         & 21 & 0    & 26 & 0    \\
\multirow{2}{*}{FastV+Random}               & 3  & 0.41 & 2  & 0.06 \\
                         & 20 & 0.06 & 20 & 0.03 \\
\shline
\end{tabular}
\end{table*}

\section{LLM Usage Statement}
We confirm that no large language models (LLMs) were used at any stage of this work, including research ideation, experimental design, data analysis, or manuscript writing. All ideas, methods, and results were conceived and executed entirely by the authors, who take full responsibility for the content of this paper.

\section{Results of Random pruning}
\label{app:token_randomvssota}

As shown in \cref{fig:random_vs_baselines_qwen_supply}, we employ various token pruning methods at different layers of the Qwen2.5-VL-7B model. Under two pruning ratios, none of the evaluated pruning methods show better performance than random pruning in deep layers across two benchmarks~\citep{fu2024mmecomprehensiveevaluationbenchmark,singh2019towards}. For the MME dataset, existing methods begin to perform no better than random pruning from the 21st layer onward. In contrast, on the TextVQA dataset, this phenomenon occurs from the 24th layer, suggesting that the model utilizes deeper layer's visual tokens for visually intensive task.

\begin{figure*}[t]
\centering
\includegraphics[width=\textwidth]{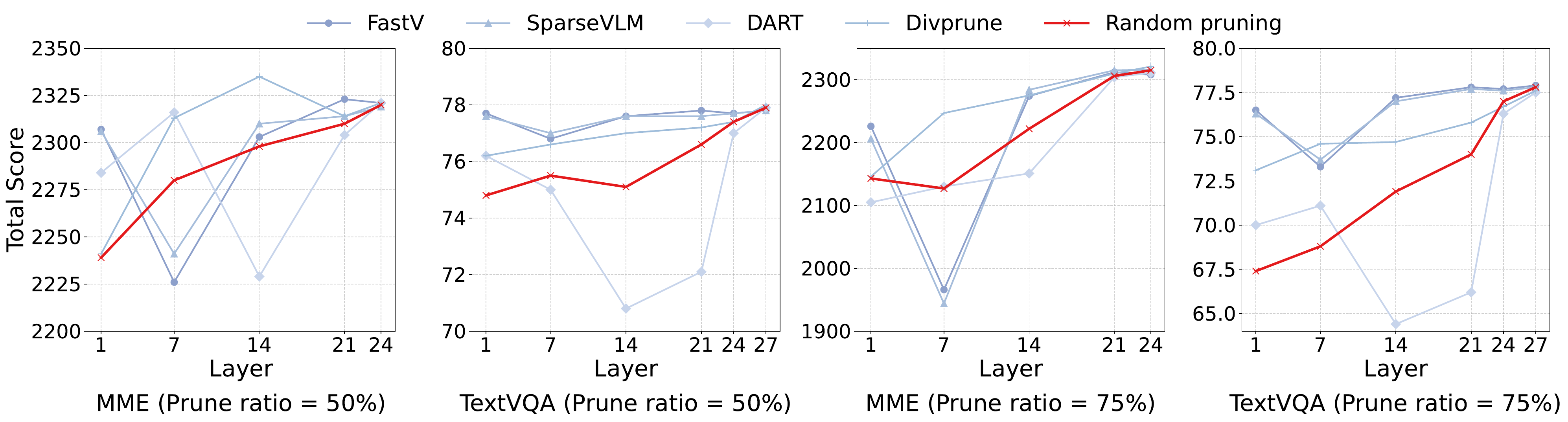}
\caption{
    Comparison of random pruning and existing methods at different pruning ratios on Qwen2.5-VL-7B.
}
\label{fig:random_vs_baselines_qwen_supply}
\end{figure*}

\section{Effectiveness of Visual Token Information on Qwen-2.5-VL-7B}
\label{app:information_effectiveness_qwen}

To verify whether the observed behavior on LLaVA-1.5-7B generalizes to other VLLMs, we further conduct our information-based pruning experiment on Qwen-2.5-VL-7B~\citep{bai2025qwen2}. Unlike LLaVA-1.5, Qwen-2.5-VL produces a varying number of visual tokens, depending on the image resolution. Following the same setup as in \cref{sec:information_effectiveness}, we compute the information value of each visual token at different decoder layers and prune low information visual tokens. Our experiments are conducted on MME~\citep{fu2024mmecomprehensiveevaluationbenchmark} and TextVQA~\citep{singh2019towards}, with four pruning ratios: 25\%, 50\%, 75\% and 87.5\%.

As shown in \cref{fig:imp_prune_qwen}, our method approaches or even outperforms the original model without pruning across all four pruning ratios evaluated on MME, consistently over all layers. It surpasses the original model performance under all four ratios after the 21st layer on TextVQA. These results indicate that our method effectively quantifies the information of each visual token on Qwen-2.5-VL as well.
Moreover, similar to the observations on LLaVA-1.5-7B, removing low-information visual tokens can lead to improved model performance.

\begin{figure*}[t]
\centering
\includegraphics[width=\textwidth]{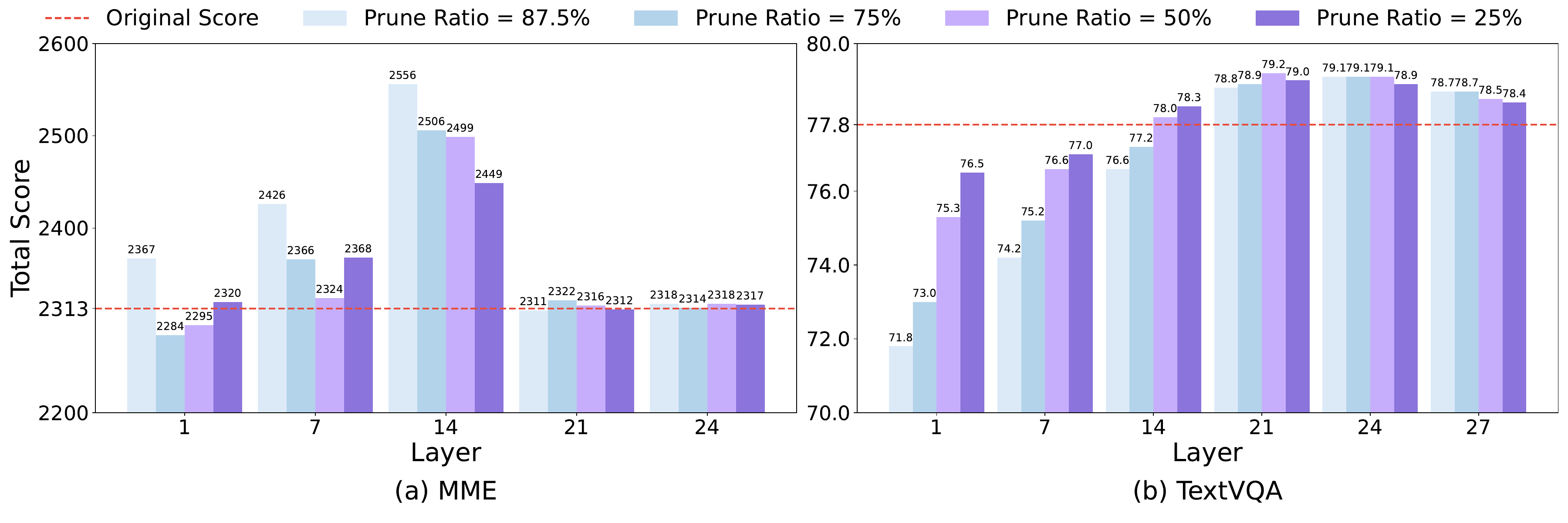}
\vspace{-2.5em}
\caption{
    \textbf{Effectiveness of proposed visual token information modeling.}
    Performance of Qwen-2.5-VL-7B on MME and TextVQA when pruning 87.5\%, 75\%, 50\% and 25\% 
    visual tokens with low information at different layers.
}
\label{fig:imp_prune_qwen}
\end{figure*}

\begin{figure*}[t]
\centering
\includegraphics[width=\textwidth]{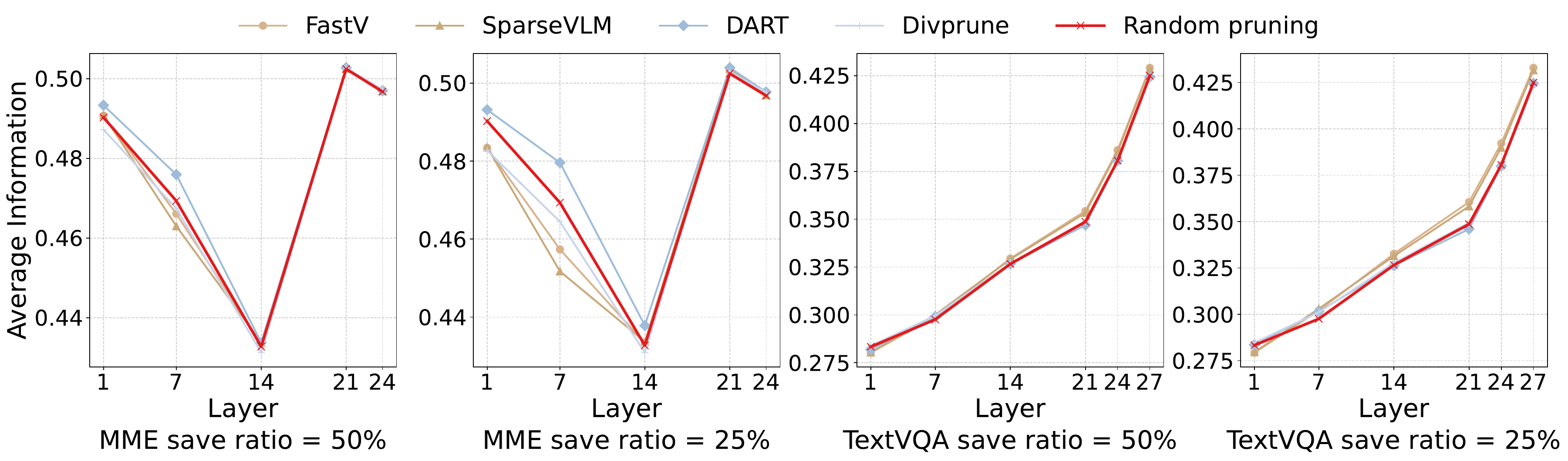}
\caption{\textbf{Evaluation of various pruning methods on Qwen-2.5-VL-7B.} We measure the sum of information in retained visual tokens when using different pruning methods.
In the deep layers, existing pruning methods fail to retain more informative visual tokens than random pruning.}
\label{fig:eval_qwen}
\end{figure*}

\section{Evaluation results of Qwen-2.5-VL-7B}
\label{app:eval_qwen}

To understand whether current pruning methods perform no better than random pruning in the deep layers, we replicate the information-preservation evaluation on Qwen-2.5-VL-7B. Specifically, we conduct experiments on both MME and TextVQA, using 1000 randomly selected samples from each dataset. We adopt two pruning ratios: 50\% and 75\%, corresponding to saving 50\% and 25\% visual tokens, respectively.

The evaluation results are illustrated in \cref{fig:eval_qwen}. On MME, the gap between existing pruning methods and random pruning begins to diminish from the 7th layer. By the 21th layer, their results become almost identical, and this pattern remains consistent in the subsequent layers. This observation aligns with the accuracy trends, as shown in \cref{fig:random_vs_baselines_qwen_supply}. For TextVQA, the gap begins to narrow at the 21st layer and becomes nearly indistinguishable by the 27th layer. As discussed in \cref{sec:When_deep}, visual tokens in TextVQA retain information at deeper layers compared with MME. This explains why the gap between random pruning and existing methods diminishes at a deeper layer on TextVQA than on MME.

\begin{figure}[t]
    \centering
    \includegraphics[width=\linewidth]{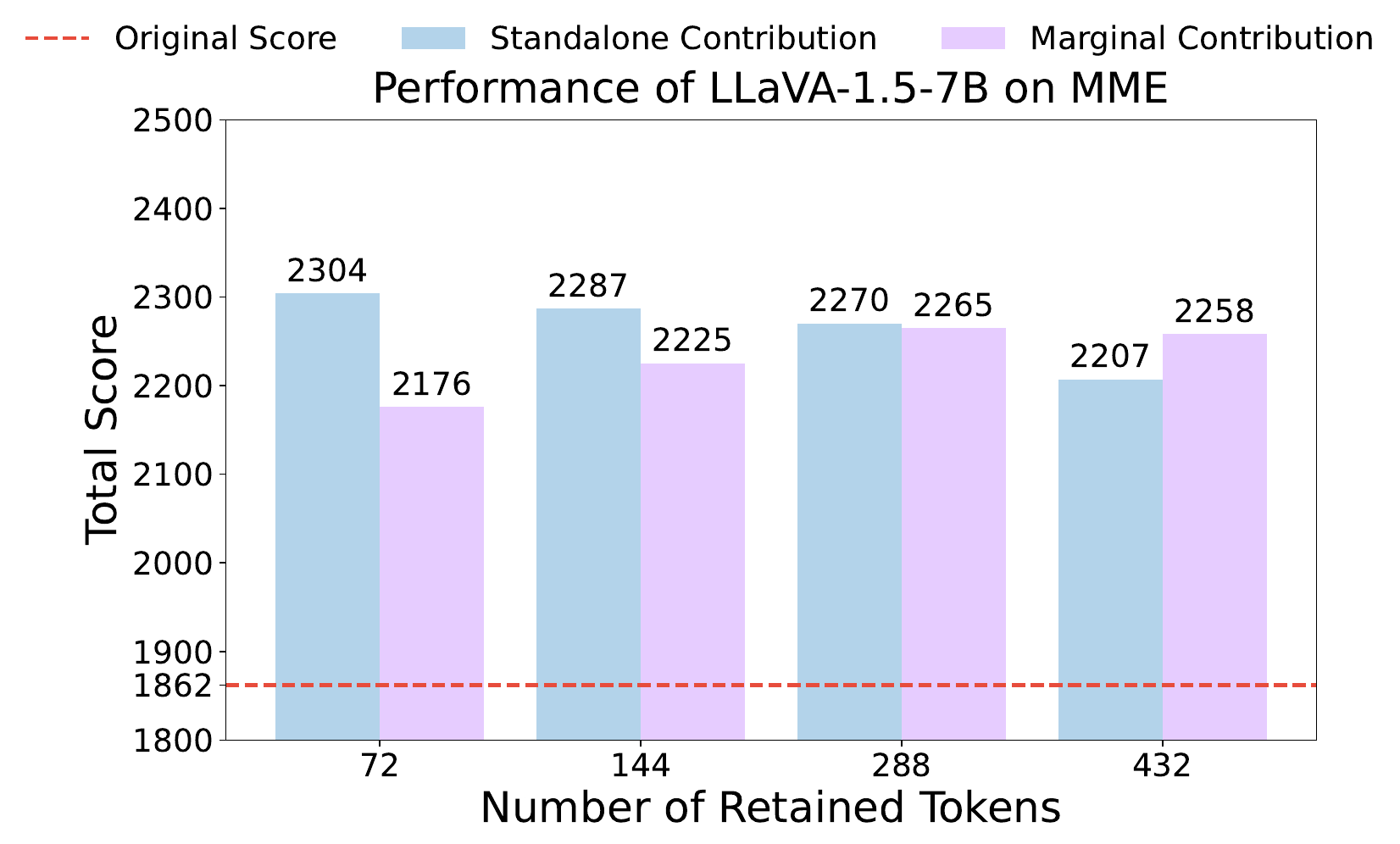}
    \caption{\textbf{Comparison of Token Information Metrics.}
    }
    \vspace{-3mm}
    \label{fig:margin}
\end{figure}

\begin{figure*}[t]
    \centering
    \includegraphics[width=\linewidth]{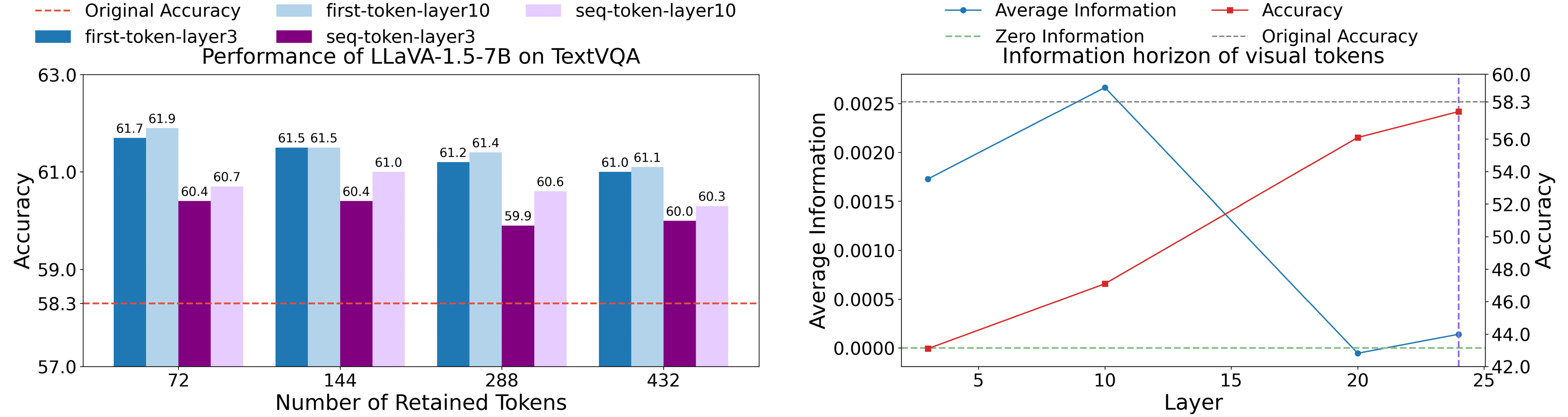}
    \caption{\textbf{Sequence-Level \textit{vs.} First-Token Metrics.}
    Left: Performance comparison on TextVQA. 
    Right: The "Information Horizon" is observed when using sequence-level metric.
    }
    \vspace{-3mm}
    \label{fig:seq-level}
\end{figure*}

\section{Justification of the Token Information Metric}
\label{app:info_metric}

Our proposed token information metric, relies on the first-token prediction probability of a single visual token against a text-only baseline. To further validate this design, we analyze two alternative definitions: Marginal-Contribution and Sequence-level metrics. 
First, we compare our Standalone-Contribution metric against the Marginal-Contribution metric (\textit{i.e.}, the probability change when removing the target token from the full set of visual tokens). As shown in \cref{fig:margin}, when pruning at the third layer of LLaVA-1.5-7B, our metric consistently matches or outperforms the Marginal-Contribution metric, particularly at high pruning ratios. 
We attribute this to visual redundancy: removing a single token often yields negligible probability changes, making it difficult to effectively rank token information. 
Second, we extend our definition to the Sequence-level, which calculates the average probability of ground-truth tokens across the entire generated sequence. While this sequence-level metric also outperforms the text-only baseline on TextVQA, the first-token metric yields superior pruning results (\cref{fig:seq-level} Left). 
We attribute this to the fact that sequence metrics may be noisy due to error propagation from earlier tokens. 
Crucially, the observed Information Horizon persists under the Sequence-level metric (\cref{fig:seq-level} Right). The average sequence information (blue line) drops in deeper layers, mirroring the behavior of the first-token metric and confirming that the information horizon is an intrinsic property of VLLMs, independent of the selection of the sequence-level or first-token metric.

\begin{figure}[t]
\centering
\includegraphics[width=\linewidth]{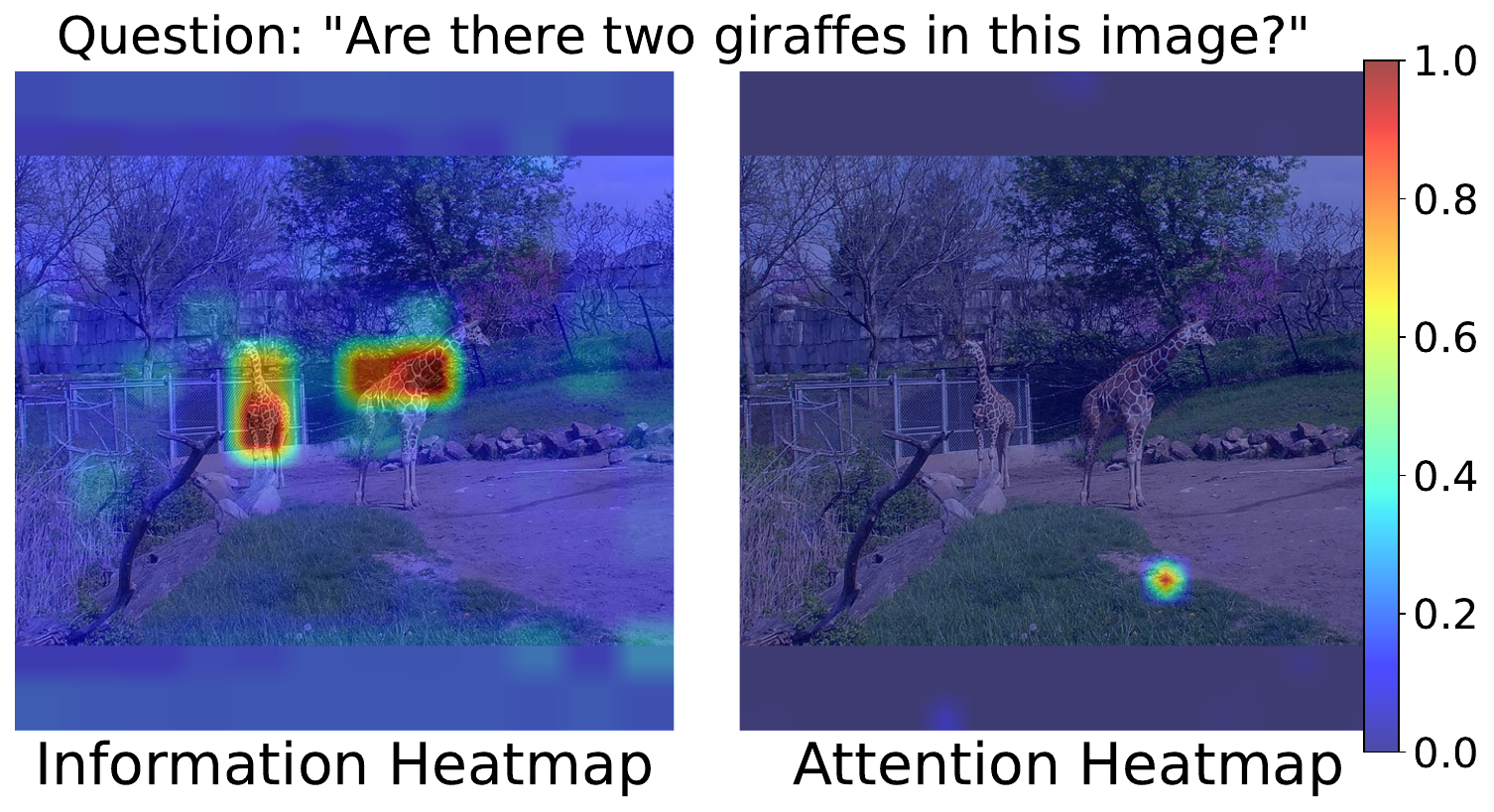}
\caption{\textbf{Visualization of information \textit{vs.} attention heatmaps.}
}
\label{fig:info_attn_map}
\end{figure}

\section{Mechanism of Low-Information Token Interference}
\label{app:mechanism_inferfence}

We attribute the performance degradation to \textbf{attention distraction}: model assigns attention weights to low-information tokens, inevitably diluting its focus on task-critical regions. 
As shown in \cref{fig:info_attn_map}, while our Information Heatmap (Left) correctly identifies the giraffes, the model's attention (Right) is \textbf{hijacked} by the irrelevant grassy background. 
Pruning these distractors forces the model to re-concentrate its attention, thereby correcting the model output. 

\begin{figure}[t]
\centering
\includegraphics[width=0.968\linewidth]{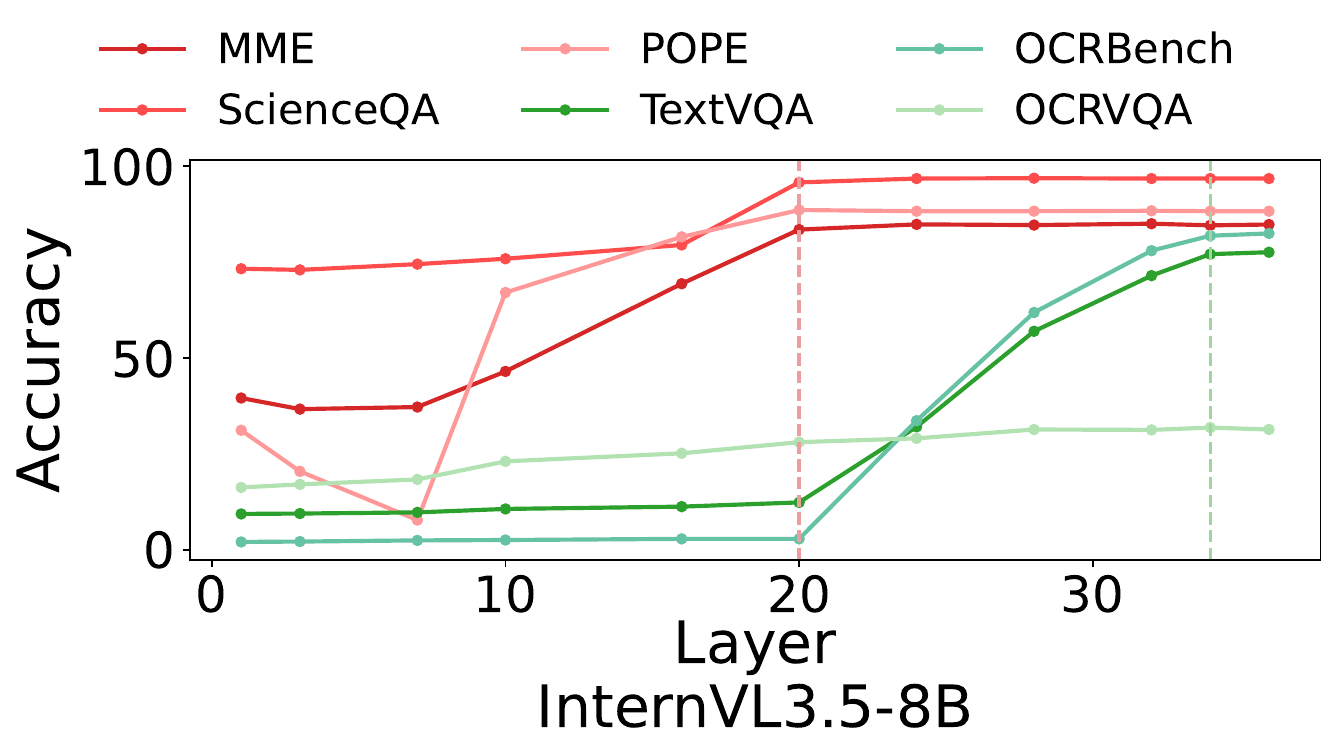}
\caption{
\textbf{Information horizon on Intern3.5VL-8B.}
}
\label{fig:internvl}
\end{figure}

\section{Generalization of Information Horizon across VLLM Architectures}
\label{app:internvl}

We validate generalizability by extending experiments to InternVL3.5-8B~\cite{wang2025internvl3_5}, with a dynamic high-resolution tokenizer and a vision encoder distinct from those used in LLaVA and Qwen.
As shown in \cref{fig:internvl}, the information horizon persists regardless of VLLM architectures: general tasks (MME, ScienceQA and POPE) saturate early (Layer 20), whereas visually intensive tasks (TextVQA, OCRBench) leverage information up to deeper layers (Layer 34), illustrating the robustness of the information horizon across varying tokenizers and model architectures.

\end{document}